\documentclass[5p,times]{elsarticle}
\usepackage[T1]{fontenc}
\usepackage{algorithmic}
\usepackage{csvsimple}
\usepackage[ruled,vlined,linesnumbered]{algorithm2e}
\usepackage{adjustbox,lipsum}
\usepackage{amsmath,amssymb,amsfonts}
\usepackage{subcaption}
\usepackage{graphicx}
\usepackage{hyperref}
\usepackage{url}
\usepackage{float}
\usepackage{ulem}
\usepackage{multirow}
\restylefloat{figure}
\usepackage[usenames]{color}

\usepackage{ulem}

\usepackage[colorinlistoftodos]{todonotes} %

\definecolor{lightgray}{HTML}{E8E8E8}

\journal{Knowledge-Based Systems}
\newdefinition{remark}{Remark}
\newdefinition{definition}{Definition}
\usetikzlibrary{trees}
\usetikzlibrary{patterns, positioning}
\usetikzlibrary{shapes.geometric, arrows}
\tikzstyle{io} = [rectangle, rounded corners, minimum width=2cm, minimum height=1cm,text centered, draw=black, fill=red!30]
\tikzstyle{estimate} = [trapezium, trapezium left angle=70, trapezium right angle=110, minimum width=2cm, minimum height=1cm, text centered, draw=black, fill=blue!30]
\tikzstyle{process} = [rectangle, minimum width=2cm, minimum height=1cm, text centered, draw=black, fill=orange!30]
\tikzstyle{decision} = [diamond, minimum width=2cm, minimum height=1cm, text centered, draw=black, fill=green!30]
\tikzstyle{arrow} = [thick,->,>=stealth]
\SetKwInput{KwInput}{Input}
\SetKwInput{KwOutput}{Output}
\begin{document}

\begin{frontmatter}

\title{
Towards a more efficient computation of individual attribute and policy contribution %
for post-hoc explanation of cooperative multi-agent systems using Myerson values
}

\author[addressaniti,addressisae]{Giorgio Angelotti}
\corref{mycorrespondingauthor}
\cortext[mycorrespondingauthor]{Corresponding author: giorgio.angelotti@isae-supaero.fr} %

\address[addressaniti]{Artificial and Natural Intelligence Toulouse Institute (ANITI), University of Toulouse, France}
\address[addressisae]{ISAE-Supaero, University of Toulouse, France}

\author[addressgranada]{Natalia Díaz-Rodríguez}

\address[addressgranada]{
\textit{Computer Science and Artificial Intelligence Department. Andalusian Research Institute in Data Science and Computational Intelligence (DaSCI) \& Research Center for Information and Communications Technologies of the University of Granada (CITIC-UGR), Spain}}
 
\begin{abstract}
A quantitative assessment of the global importance of an agent in a team is as valuable as gold for strategists, decision-makers, and sports coaches. Yet, retrieving this information is not trivial since in a cooperative task it is hard to isolate the performance of an individual from the one of the whole team. Moreover, it is not always clear the relationship between the role of an agent and his personal attributes. In this work we conceive an application of the Shapley analysis for studying the contribution of both agent policies and attributes, putting them on equal footing. Since the computational complexity is NP-hard and scales exponentially with the number of participants in a transferable utility coalitional game, we resort to exploiting a-priori knowledge about the rules of the game to constrain the relations between the participants over a graph.
We hence propose a method to determine a Hierarchical Knowledge Graph of agents' policies and features in a Multi-Agent System.
Assuming a simulator of the system is available, the graph structure allows to exploit dynamic programming to assess the importances in a much faster way.
We test the proposed approach in a proof-of-case environment deploying both hardcoded policies and policies obtained via Deep Reinforcement Learning. The proposed paradigm is less computationally demanding than trivially computing the Shapley values and provides great insight not only into the importance of an agent in a team but also into the attributes needed to deploy the policy at its best.
\end{abstract}

\end{frontmatter}

\section{Introduction}

Research in the field of Multi-Agent Systems (MAS) suggests viable pathways to solve complex tasks \cite{dorri2018multi}. In a MAS environment, every agent is, in principle, an individual independent of one another with its own characteristics and skills. The main idea is that by assigning to each agent a specific subtask according to its perks and hence exploiting a delocalized control, it is possible to solve a problem more efficiently. The human society itself is an example of a MAS since groups of individuals usually train according to their nature to exercise specific professions that require different expertise: medical personnel, firefighters, engineers, etc.
When analyzing the behavior of agents in a MAS a question arises immediately: according to a common goal to be reached, which agent is contributing the most, and which are its most important individual attributes?
There is not a trivial answer to such inquiry, since it is hard to assess quantitatively the quality of the interaction and cooperation among agents.

The branch of mathematics that studies the cooperation between agents towards a common goal is called cooperative game theory. In 1951 Shapley formalized a paradigm to assess the contribution of single agents in what is called a transferable utility coalitional game \cite{aumann2015values}.
While Shapley's analysis was originally thought to quantify the worth of human agents in a team, its application is straightforward to every other possible transferable utility coalitional game that respects the needed mathematical properties.

The field of possible applications of Shapley and Myerson analyses or their generalizations is broad. Shapley analysis or its suitable generalizations can be applied for instance to estimate the contributions of basketball players in a match using the recorded match data and statistics \cite{metulini2022measuring}. If the practitioner possesses some information about the connectivity of interactions, or, e.g., spatial rules of the game that restrict the interaction among agents, Shapley and Myerson analyses can be used to assess the importance of vertices, i.e., agents, in graphs. Recent works investigated the Shapley and Myerson analyses of transportation networks \cite{hadas2017approach} and bus-holding strategies \cite{dai2019predictive}. Cooperative game theory can be applied also to more abstract tasks spanning from the assessment of feature contribution in a classification/regression task solved through machine learning models \cite{lundberg2017unified} to a purpose much closer to our fundamental objective: the explainability of opaque models through the contribution of agents in Cooperative Multi-Agent Reinforcement Learning (RL) \cite{heuillet2022collective, wang2020cooperative} %
and MAS \cite{moya2017agent,moya2021simulating,giraldez2020modeling,fisher2022beaut}.

\subsection{Explainable Artificial Intelligence (XAI)}
Explainable Artificial Intelligence (XAI) \cite{gunning2019xai} is coming more and more %
ubiquitously on the agenda of researchers since the eventual application of algorithms and black-box methods in the real world concerns both general public trust, policy- and lawmakers.
Current XAI methods encompass the explainability of regression and classification models, but little work has been done about the explainability of behavioral policy and MASs, and more concretely in the realm of reinforcement learning.

State-of-the-art eXplainable Reinforcement Learning (XRL) methods \cite{heuillet2021explainability} can be split into two branches: interpretable by design or transparent %
methods and post-hoc explainability techniques. These all can be found in XAI taxonomies \cite{arrieta2020explainable, heuillet2021explainability}. Transparent models %
can be directly explained since the model's interpretability is inherent. These are composed of a sequence of straightforward understandable functions, for instance, decision trees.
Not only the output of the technique is explainable but also the computation flow itself and from here the name \textit{transparent}.

Post-hoc techniques encompass the other large family of methods that develop a (usually non-trivial) posterior interpretation of the output. A large number of %
XRL baselines exploit %
(post-hoc) techniques \cite{heuillet2021explainability}, while few others make use of the explanation to modify the training process. One such example is in \cite{portugal2022analysis}, where Deep Neural Networks are used to adapt Explainable Goal-Driven RL to continuous environments. They try to develop goals that are also explainable during the training phase of a single agent and exploit the contribution-based explanations to accelerate the convergence of the learning algorithm towards a better performance policy. This technique is not post-hoc but rather online and little can tell about the quality of the policy in a MAS. 

On the post-hoc family of methods, one of the most common model-agnostic post-hoc XAI techniques is analyzing machine learning (ML) model outputs via SHapley Additive exPlanations (SHAP) \cite{lundberg2017unified}. SHAP is an additive feature attribution framework based on game theory that fairly assigns the won payout in a cooperative game between players by additively decomposing, as a linear explanation model $g$, the prediction (or \textit{credit}) of a (to be explained) ML model $f$ among all features involved in the prediction. The work in \cite{lundberg2017unified} proved that the only per-feature linear coefficients that guarantee the three necessary properties of a good explanation model (local accuracy, missingness, and consistency) coincide with the Shapley values of the coalitional game \textit{played} by the input features.

Since computing Shapley values is an NP-hard problem \cite{michalak2013efficient} whose complexity scales exponentially in the number of players (features), SHAP \cite{lundberg2017unified} speeds up this process by approximating the Shapley value using a so-called SHAP Kernel obtained through weighted linear regression and an ad hoc sampling method. %

Unfortunately, despite SHAP's great success in explaining models used for regression and classification, the approach is not directly applicable to RL problems because in RL the notion of data point, necessary for the execution of the method, does not exist \textit{per se}.

Going beyond the tasks of regression and classification, the recent work in \cite{heuillet2022collective} developed an approach to directly estimate the importance of agents in cooperative multi-agent RL environments resorting to a Monte-Carlo estimation of Shapley values \cite{aumann2015values}. %

The results in this study are not surprising since Shapley values were originally defined to compute the importance values of players in generic \textit{transferable utility coalitional} games \cite{Peters2008}, i.e., games where players have to cooperate to gain a common outcome and, hence, to redistribute the spoils (or \textit{payout}) in a worth-based proportional fashion. 
 While the work in \cite{heuillet2022collective} produces reasonable predictions of individual agents' policy importance, not much is %
possible to elucidate why a policy is more important than another and what individual statistics are required to make the said policy work.

In this manuscript, we extend the work of \cite{heuillet2022collective} to take into account not only the policy of individual agents but also their individual attributes. For instance, if we were to analyze a football game, we would not only evaluate the policy of defenders, forwarders and midfielders, etc., but also their individual skills: pass accuracy, speed, stamina, etc.

This kind of analysis can produce valuable insights since it can happen that poor resource management results in allocating a task (assigning a role) to users that are not suited to carry it out.
Consider the extreme example of a football manager that deploys a very strong defender with poor shooting accuracy as the only forward of his team. Even if the agent is strong he will not be important to the team since his policy (being a forward) does not exploit his perks which are indeed specific to a good defender. The latter was a very trivial example, but in real-life situations or more realistic environments, the relationships between policies and attributes could be highly non-linear and not so easily directly understandable.
With this in mind, our goal is to shed light not only on why in a cooperative game one agent is more important than another but also on explaining which of its individual skills are needed to exert his role to the best of his potential.
To do so, we will consider a transferable utility coalitional game where the \textit{players} are not only the individual features of each agent but also the policies. More precisely, every separately taken individual agent's policy and individual agent's feature will be a \textit{player} of a coalitional game with transferable utility, i.e., with the possible sharing of utility (a real number) across the participating players.

Recently, also the work in \cite{metulini2022measuring} proposed to analyze both agents' policies and attributes of basketball players resorting to generalized Shapley analysis. However, their work differs from ours in two main aspects: 1) the individual attributes are not participants of the coalitional game, but rather are included in the definition of the generalized characteristic function that is then queried only on coalitions of policies (agents), 2) the characteristic function is evaluated on recorded data and does not need a simulator (to run rollouts and average the values to account for the stochasticity of more realistic MAS environments).

In a transferable utility coalitional game with a larger number of participants, the computation of Shapley values is more computationally demanding, since the number of possible coalitions over the subsets of players grows faster than exponentially with the total number of participants. We will attenuate this effect by exploiting a-priori knowledge of the structure of the game and computing the Myerson values, i.e., the equivalent of Shapley values on a cooperative game constrained on a graph \cite{myerson1977graphs}. The information about the said structure in the game will allow for fewer computations, hence improving its scalability.

Just to provide some insights, in the use-case scenario we will present later we will consider $3$ agents, each one deploying its own policy and each one characterized by $4$ individual attributes. While the work in \cite{heuillet2022collective} would just analyze the $3$ policies, we will examine a total of $15$ elements: the $3$ policies and the $12$ individual attributes. However, as previously stated, the complexity of the problem is NP-hard \cite{michalak2013efficient} and the %
computational %
cost needed to solve the problem increases exponentially in the number of the analyzed elements. Precautions should be taken to obtain more computationally efficient and hence environmentally friendly methods. Indeed, less polluting algorithms became increasingly topical in recent years in the spirit of the necessary \textit{Green AI} \cite{schwartz2020green}.
Since we will focus just on the post-hoc explainability aspect of MASs we will analyze both handcrafted cooperative policies and agent's policies in a multi-agent environment, and a policy learned through Deep Reinforcement Learning. The policy learning phase is outside the scope of this paper.

This work could inspire future research not only in XAI but also provide useful insights into political (and geopolitical) analysis, considering not only the orientation of a country but also its resources. Further applications involve economics, management, fair policy design, decision-making, sports, and, more generally, social dilemmas. 
\begin{figure*}[t]%
\centering
    \begin{subfigure}{\textwidth}
    \centering
    \begin{tikzpicture}[node distance=4.2cm]
    
    \node (start) [estimate,text width=3cm] {List policies and individual features of each agent};
    \node (rep) [process, right of=start,text width=3cm] {Define replacement rules};
    \node (shap) [io, right of=rep,text width=4cm] {Run Shapley analysis (Algorithm \ref{algo:shapley-sim})};
    \draw [arrow] (start) -- (rep);
    \draw [arrow] (rep) -- (shap);
    \end{tikzpicture}
    \caption{Flowchart of the methodology with a simulator and Shapley analysis.}\label{fig:flow-shap}
        
    \end{subfigure}\\%
    \begin{subfigure}{\textwidth}
    \centering
        \begin{tikzpicture}[node distance=4.2cm]
    
    \node (start) [estimate,text width=3cm] {List policies and individual features of each agent};
    \node (rep) [process, right of=start,text width=3cm] {Define replacement rules};
    \node (hkg) [decision, right of=rep,text width=1cm] {Build HKG};
    \node (mye) [io, right of=hkg,text width=4cm] {Run Myerson analysis (Algorithm \ref{algo:myerson-sim})};
    \draw [arrow] (start) -- (rep);
    \draw [arrow] (rep) -- (hkg);
    \draw [arrow] (hkg) -- (mye);
    \end{tikzpicture}
    \caption{Flowchart of the methodology with a simulator and Myerson analysis.}\label{fig:flow-mye}
    \end{subfigure}
    \caption{Flowcharts of the proposed methodologies.}\label{fig:flow}
\end{figure*}
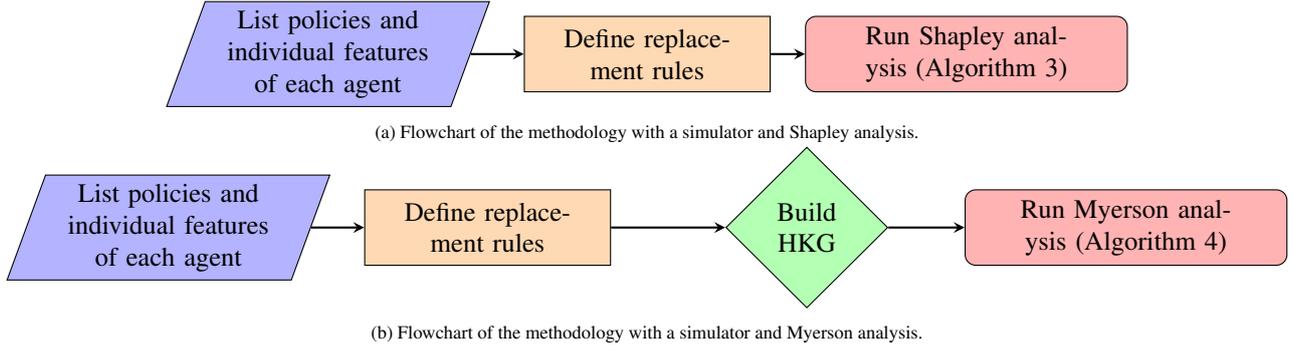
The paper is structured as follows:
\begin{enumerate}
    \item In Section \ref{seq:def} the mathematical concepts needed to set the base to propose our approach are briefly introduced;
    \item In Section \ref{seq:tree} we present the first contribution: a protocol to build a Hierarchical Knowledge Graph (HKG) for Multi-Agent Systems is proposed;
    \item Section \ref{seq:sim} lists the computational precautions that a practitioner should take to compute Shapley or Myerson values for Multi-Agent Systems using a simulator: \textit{e.g.}, how to compute the characteristic function at a coalition without the full number of features or policies: e.g., in a system of three agents A, B and C, a coalition made of only the policy of A and the policy of B, thus missing the policy of C.
    \item The main experimental contributions are presented in Section \ref{seq:exp}:
    \begin{enumerate}
        \item Designing a transferable utility coalitional game to assess the importance of agents' policies and features in a multi-agent cooperative environment in which the \textit{participants} are represented by \textit{both} the agents' policies and their individual attributes;
        \item Exploiting a-priori knowledge of the game to constrain the computation of Myerson values on a graph;
        \item Testing the approach in a small yet representative environment;
    \end{enumerate}
    \item Results are discussed in Section \ref{seq:discussion};
    \item Further experiments are performed deploying a policy learned by a Deep RL black-box model in Section \ref{sec:xai};
    \item Finally, conclusions and future perspectives are highlighted in Section \ref{seq:conclusions}.
\end{enumerate}

\paragraph{Contributions of this work}
This manuscript proposes a methodology to explain a Multi-Agent System by assessing the importance of factors that contribute to the achievement of a common goal. The methodology consists of:
\begin{enumerate}
    \item The proposal of aligning both agents' policies and individual attributes on equal footing with the aim to assess their importance with respect to the goal of the Multi-Agent System using Shapley and Myerson analyses and a simulator (establishing suitable replacement rules);
    \item The introduction of an expert-guided protocol to build a knowledge graph depicting the connectivity in the interactions between agents' policies and individual attributes in a Multi-Agent System. This graph will be defined as a Hierarchical Knowledge Graph;
    \item The empirical validation of (1) in a use-case Multi-Agent System using a simulator, showing that the computed Shapley values are consistent with the system dynamics and that relevant features can be detected using this approach;
    \item Exploiting (2) for the empirical validation of (1), showing that the protocol to build a Hierarchical Knowledge Graph captures the connectivity of interactions in the system since the consequent Myerson analysis is statistically equivalent to the Shapley one and, on top of that, less computationally expensive by an amount that depends on the case-specific graph structure;
    \item Empirically showing that the operationalized technique can provide explanations regarding the accountability and roles of policies and attributes of agents trained with black box models such as Deep Reinforcement Learning architectures.
\end{enumerate}
Comprehensive flowcharts of the proposed methodologies are displayed in Figure \ref{fig:flow}.

\section{Formal Background}
\label{seq:def}

In this section, we define transferable utility coalitional games, Shapley and Myerson values. These two quantities are the main characters of the present work and need to be thoroughly introduced.
\begin{definition}\label{def:coal}[Transferable utility coalitional game \cite{Peters2008}]
Let $\mathcal{C}$ be a finite set of players ($\lvert\mathcal{C}\rvert \in \mathbb{N}_{+}$) and let characteristic function $v: \mathcal{P}\left(\mathcal{C}\right) \rightarrow \mathbb{R}$ with $\mathcal{P}\left(\mathcal{C}\right)$ being the power set of $\mathcal{C}$, i.e., a coalition of features.  $v$ is called the characteristic function and maps subsets of players (coalitions) to real numbers. 

\textit{Let $v$ be endowed of the following property:}
\begin{equation}
    v(\varnothing) = 0.
\end{equation}
A transferable utility coalitional game $G$ is defined as the tuple $G = \left(\mathcal{C}, v\right)$.
\label{eq:null}
\end{definition}

\begin{remark}
The characteristic function $v$ describes the worth (utility) of a coalition of players in the game when they cooperate. The worth of an empty coalition is zero.
\end{remark}
\begin{remark}
The word player is used just to provide intuition and coherence with Shapley's game theory background \cite{molnar2020interpretable}. A member of the set of players $\mathcal{C}$ could represent anything, and indeed later on in the manuscript, it will be composed of individual agent's attributes and policies.
\end{remark}
\begin{definition}[Shapley value \cite{aumann2015values}]Shapley analysis \cite{lundberg2017unified} allows to compute the Shapley value of a player $i \in \mathcal{C}$ in a transferable utility coalitional game $G = \left(\mathcal{C}, v\right)$, and it is defined as:
\begin{equation}
\label{eq:shapley}
    \phi_{i}(v) = \sum_{K \subseteq \mathcal{C}\backslash\{i\}}\frac{\lvert K\rvert! \left(\lvert \mathcal{C}\rvert - \lvert K\rvert -1\right)!}{\lvert \mathcal{C}\rvert!} \left(v\left(K\cup \{i\}\right) - v\left(K\right)\right).
\end{equation}
\end{definition}
\begin{remark}
The Shapley value assigns to every player a real number corresponding to its importance in the game. It is computed by averaging the difference in the worth of every possible coalition with and without the player.
\end{remark}
The Shapley value redistributes the worth of players abiding by the following properties:
\begin{enumerate}
    \item \textit{Efficiency}: The contribution of each player in the transferable utility coalitional game adds up to the characteristic value $v$ computed on the coalition including all players
    \begin{equation}\sum_{i \in \mathcal{C}}\phi_{i}(v) = v(\mathcal{C});\end{equation}
    
    \item \textit{Symmetry}: If adding feature element $i$ or the element $j$ to any coalition that originally does not include those elements results in the same evaluation of the characteristic function, then elements (in our use case \textit{features} or \textit{policies}) $i$ and $j$ contribute the same.
    \begin{gather}
        \nonumber\text{if}\ v(K\cup\{i\}) = v(K\cup\{j\})\   \forall\mkern2mu K\subseteq\mathcal{C} : \{i,j\}\subsetneq K \\
        \implies \phi_{i}(v) = \phi_{j}(v);\end{gather}
    \item \textit{Linearity}: The contribution of a player to two transferable utility coalitional games played by the same team, but with different characteristic functions $v$ and $w$, can be linearly composed.
    \begin{gather}
    \text{Let } G_{1} = (\mathcal{C}, v) \land G_{2} = (\mathcal{C}, w) \\
    \nonumber\implies \phi_{i}(av+bw) = a\phi_{i}(v) + b\phi_{i}(w),  \forall\mkern2mu i \in \mathcal{C},\  \forall\mkern2mu (a,b) \in \mathbb{R}^{2};
    \end{gather}
    Note that linearity is defined over characteristic function $v$, and it is different from the Decomposition property of graph-constrained transferable utility coalitional games (to be defined in Equation \ref{eq:grapheq}), since the latter allows to linearly decompose the computation of $v$ over a coalition into the linear combination \textit{of the same $v$} evaluated over the connected components of the graph.
    \item \textit{Null player}: if adding a player $i$ to every coalition that did not have it does not affect the computation of the characteristic function, then the contribution of element $i$ is zero.
    \begin{gather}
    \label{eq:nplayer}
    \nonumber\text{If } v(K\cup\{i\}) = v(K)\    \forall\mkern2mu K \subseteq\  \mathcal{C} \backslash \{i\} \\
    \implies \text{player $i$ is said to be \textit{null}} \land \phi_{i}(v)=0.\end{gather}
\end{enumerate}

In summary, $\phi_{i}(v)$ is the Shapley value of element $i\in \mathcal{C}$ in the transferable utility coalitional game defined by $(\mathcal{C}, v)$. 
$v$ is the characteristic function: a function that maps coalitions of elements of the set $\mathcal{C}$ to real numbers that represent the payoff or return of the game in the context of Shapley analysis.
\begin{figure}
    \centering
    \includegraphics[width=\columnwidth]{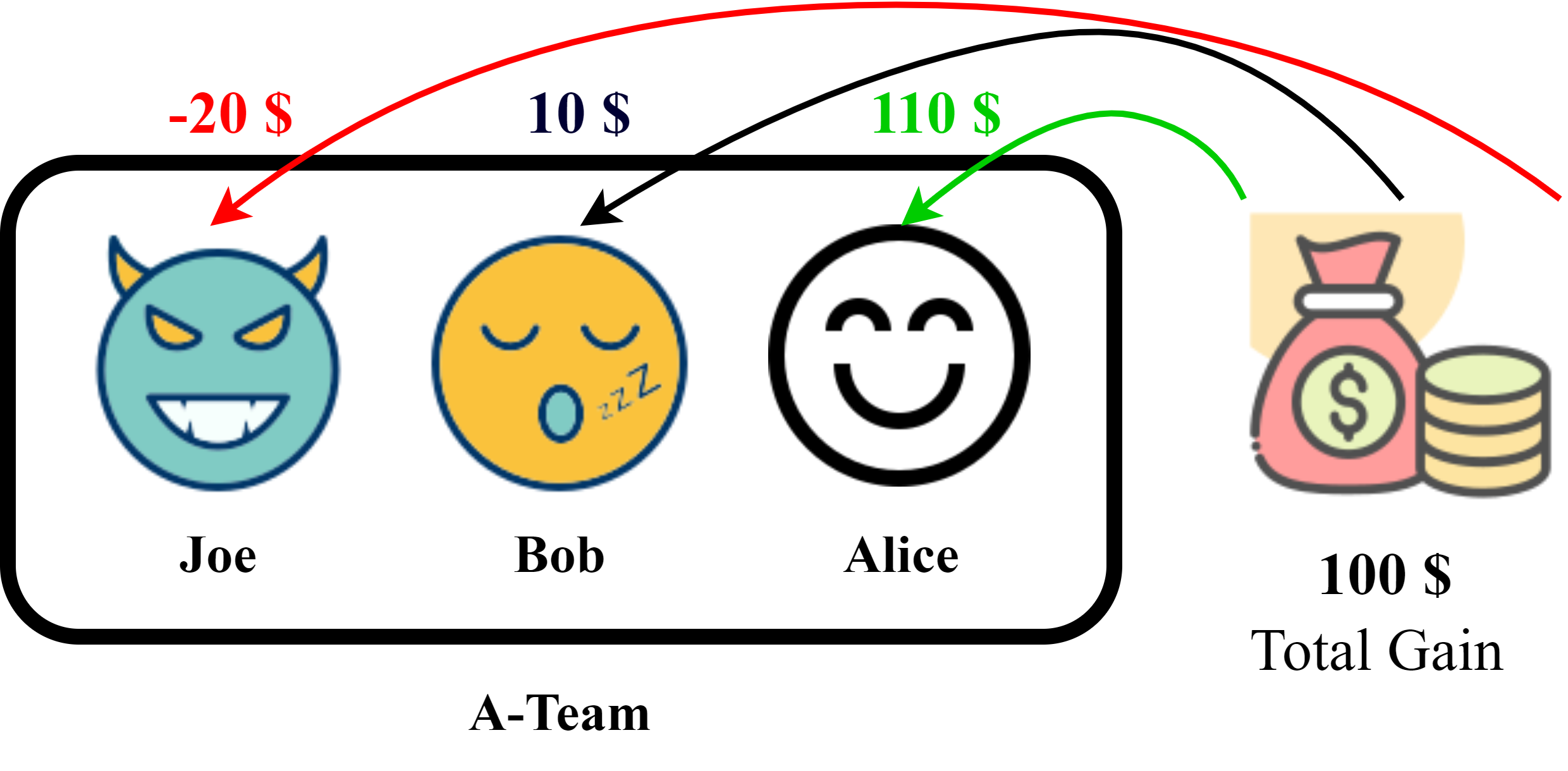}
    \caption{Example: a group of computer engineers called the A-Team is paid $100$ \$ for a project. The members of the group are Alice, a very diligent and happy worker, Bob, who is always sleepy and unproductive, and Joe, an incognito saboteur working for a competing company. A Shapley analysis would find, for instance, a contribution of $110$ \$ for Alice, $10$ \$ for Bob, and $-20$ \$ for Joe. Joe's deeds are detrimental to the project and in an ideal world, he should refund the others.}
    \label{fig:ateam}
\end{figure}
Shapley developed this tool to describe the contribution of real (human) team members trying to reach a goal represented by a quantifiable economic benefit. Intuitively, one can think of $v$ as how much payoff a subset of "players" (of no matter what size) earns/loses if they cooperate (see Figure \ref{fig:ateam} for an example).
In our application, the requirement in Equation \ref{eq:null} is necessary to allow the explanation of the importance of both \textit{policies} and \textit{attributes}.
The algorithm to calculate Shapley values of players (policies/attributes) is reported as Algorithm \ref{algo:shapley} (Line 5 is Equation \ref{eq:shapley}).
\begin{algorithm}
\caption{Exact Shapley Values Computation}\label{algo:shapley}
\KwInput{$v$ characteristic function of the coalitional game $(\mathcal{C},v)$}
\KwOutput{$i=1$ \KwTo $\lvert\mathcal{C}\rvert$ values $\phi_i(v)$ Shapley values}
 \textbf{Initialization:} $\phi_{i}(v)= 0$ $\forall i \in \mathcal{C}$\\
 \For{$i \in \mathcal{C}$}{
 Generate the power set $\mathcal{P}\left(\mathcal{C}\backslash\{i\}\right)$\\
 \For{$K \in \mathcal{P}\left(\mathcal{C}\backslash\{i\}\right)$}{
 $\phi_i(v) \leftarrow \phi_i(v) + \frac{\lvert K\rvert! \left(\lvert \mathcal{C}\rvert - \lvert K\rvert -1\right)!}{\lvert \mathcal{C}\rvert!} \left(v\left(K\cup \{i\}\right) - v\left(K\right)\right)$ }
 }
\end{algorithm}
Definition \ref{def:coal} refers to $\mathcal{C}$ as being a finite set of elements, it does not specify of what \textit{kind}. What is important is that all mathematical properties listed in Definition \ref{def:coal}, Equations \ref{eq:null}-\ref{eq:nplayer} are respected.

\begin{figure*}
    \centering
    \includegraphics[width=0.78\textwidth]{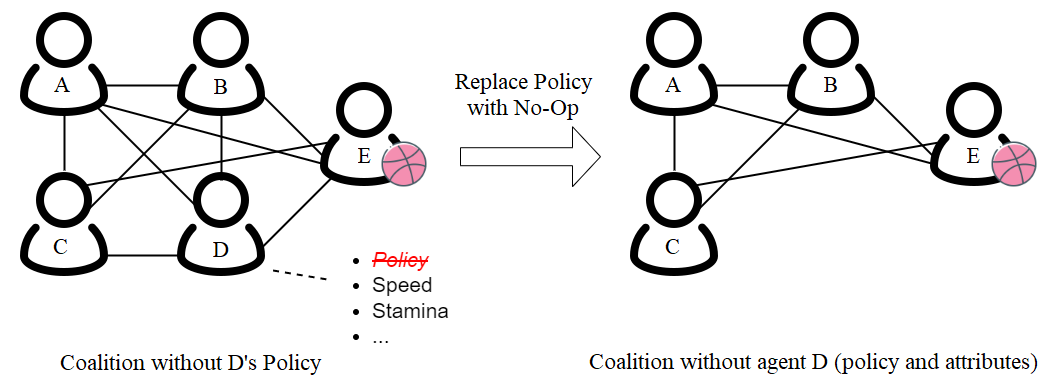}
    \caption{Feature and policy coalition formation example: Five basketball players (A, B, C, D, E) interact with each other during a match. We consider a set of policies and individual attributes (as properties) connected to each one of the agents. Both, attributes and policies are considered as players in the cooperative game coalition. When querying the characteristic function for a coalition without considering D's policy, since it would be replaced by the No-op policy for consistency, computing the characteristic function amounts to applying it to the same original team, without agent D (and %
   without the whole set of attributes that were not removed).
    \label{fig:basketball}}
\end{figure*}

For example, it is mandatory to respect Equation \ref{eq:null} and therefore, when applying the approach to a real-world scenario or to an RL environment where one can never consider a simulation where a player (policy or feature) is \textit{entirely removed}, how to deal with the fact that one wants a void coalition or a coalition without some features/policies?
It is crucial to set some rules that: 1) ease the removal of policies/features, 2) when one wants to "remove" all elements of $\mathcal{C}$ then $v(\emptyset) = 0$. In the Arena game that will be introduced in the next sections, this rule translates into replacing any policy with \textit{No-Op}, and the attribute values with $0$.
This explanation is consistent with what was found \textit{empirically} by Heuillet et al. 2022 \cite{heuillet2022collective} (i.e., the \textit{No-Op} policy of doing nothing was the best way of replacing a policy). However, the said rule is not general, it depends on how the characteristic function $v$ acts. This is why we will define in that way the score formula, in Equation \ref{eq:score}, to have $S=0$ for an empty coalition (see Figure \ref{fig:basketball} for an example).

\begin{definition}[Transferable utility coalitional game over a graph]
\textit{Let $G = \left(\mathcal{C}, v\right)$ be a transferable utility coalitional game and let $\Gamma$ be a graph that has the set of players $\mathcal{C}$ as a set of nodes. We can define the transferable utility coalitional game $G$ over the graph $\Gamma$ by limiting the interaction amongst players through the establishment of edges in the graph. The said transferable utility coalitional game will be denoted as $G_{\Gamma}$.}

\noindent\textit{Let $\mu$ be a function that maps the set of nodes $K$ of a subgraph of  $\Gamma$ to the elements of its minimum connected node cover (the cover of nodes of $\Gamma$ that induces the minimum number of connected subgraphs). %
}

\textit{%
\noindent$v$ is hence endowed of the following property in addition to Shapley values' properties 1-4 that is due to the limited interaction between participants dictated by the graph structure:}\\

\indent 5. Coalition decomposition: \textit{the characteristic function evaluated on a coalition $K$ is equal to the sum of the characteristic function evaluated over the elements of the minimum node cover of the subgraph $\Gamma$ whose nodes are in $K$: }
\begin{gather}
\label{eq:grapheq}
    \nonumber v(K) = \sum_{\sigma\in \mu(K)}v(\sigma)\\
    \forall\mkern2mu K \in \mathcal{P}(\mathcal{C}) : \bigcup_{\sigma\in \mu(K)} \sigma = K,\ 
    \left(\sigma_{i}\cap\sigma_{j} = \emptyset\right) \  \forall\mkern2mu (\sigma_{i},\sigma_{j})\in \mu^{2}(K).
\end{gather}
\end{definition}
\vspace{-0.6cm}
\begin{algorithm}
\caption{Exact Myerson Values Computation}\label{algo:myerson}
\KwInput{$\Gamma$ graph over the set of players $\mathcal{C}$ of the coalitional game $(\mathcal{C},v)$, $v$ characteristic function of the coalitional game $(\mathcal{C},v)$}
\KwOutput{$i=1$ \KwTo $\lvert\mathcal{C}\rvert$ values $\phi^{n}_i(v)$ Myerson values}
 \textbf{Initialization:} $\phi_{i}(v)= 0$ $\forall i \in \mathcal{C}$, Coalitions = $\{\varnothing\}$\\
 \For{$i \in \mathcal{C}$}{
 Generate the power set $\mathcal{P}\left(\mathcal{C}\backslash\{i\}\right)$\\
 Decompose each $K \in \mathcal{P}\left(\mathcal{C}\backslash\{i\}\right)$ to $\mu(K)$, the sets of connected nodes minimally covering the subgraph with vertices $K$\\
 Decompose each $K \cup \{i\}$ to $\mu(K \cup \{i\})$, the sets of connected nodes minimally covering the subgraph with vertices $K \cup \{i\}$\\
 \For{$K \in \mathcal{P}\left(\mathcal{C}\backslash\{i\}\right)$ ordered by increasing $\lvert K \rvert$}{
  \If{$K \not\in$ Coalitions}
  {\For{$\sigma \in \mu(K) \land \sigma \not\in\text{Coalitions}$}{Coalitions$(\sigma) \leftarrow v(\sigma)$}Coalitions$(K) \leftarrow \sum_{\sigma \in \mu(K)}\text{Coalitions}(\sigma)$}
\If{$K \cup \{i\} \not\in$ Coalitions}
  {\For{$\sigma \in \mu(K\cup\{i\}) \land \sigma \not\in\text{Coalitions}$}{Coalitions$(\sigma) \leftarrow v(\sigma)$}Coalitions$(K\cup\{i\}) \leftarrow \sum_{\sigma \in \mu(K\cup\{i\})}\text{Coalitions}(\sigma)$}
 $\phi_i(v) \leftarrow \phi_i(v) + \frac{\lvert K\rvert! \left(\lvert \mathcal{C}\rvert - \lvert K\rvert -1\right)!}{\lvert \mathcal{C}\rvert!} \left(\text{Coalitions}\left(K\cup \{i\}\right) - \text{Coalitions}\left(K\right)\right)$ }
 }
\end{algorithm}

\textit{where $\sigma$ is a dummy variable that spans the elements of the minimum node cover of $\Gamma$. Note that the elements of a node cover are sets of nodes. }\\

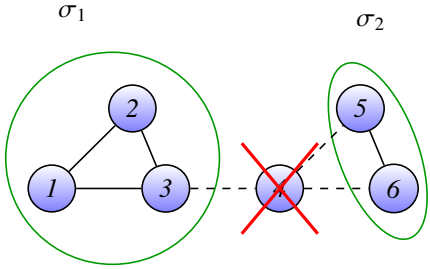
\begin{figure}[ht]
\centering
\begin {tikzpicture}[-latex, auto, node distance =1.5 cm and 1.5 cm, on grid, semithick, text=black,
node/.style ={circle, top color = white, bottom color=blue!50,
draw, black, text=black, minimum width = 0.2 cm},
labels/.style ={circle, top color = white, bottom color=white,
draw, white, text=black}]

\node[node] (1) {\textit{1}};
\node[node] (2) [above right of=1]{\textit{2}};
\node[node] (3) [right of=1]{\textit{3}};
\node[node] (4) [right of=3]{\textit{4}};
\node[node] (5) [above right of=4]{\textit{5}};
\node[node] (6) [right of=4]{\textit{6}};

\draw[black!40!green] (0.8, 0.4) circle (40pt);
\node[labels,label=$\sigma_1$] at (0.28,1.92) {};
\draw[black!40!green,rotate=20] (4.2, -1) ellipse (15 pt and 35pt);
\node[labels,label=$\sigma_2$] at (4.2,1.8) {};

\path[-] (1) edge (2);
\path[-] (1) edge (3);
\path[-] (2) edge (3);

\path[-,dashed] (3) edge (4);

\path[-,dashed] (4) edge (5);
\path[-,dashed] (4) edge (6);
\path[-] (5) edge (6);

\draw[-, red, very thick] (2.5, 0.6) -- (3.5,-0.6);
\draw[-, red, very thick] (2.5, -0.6) -- (3.5,0.6);

\end{tikzpicture}
\caption{Example of a graph $\Gamma$ over a set of nodes $\mathcal{C}=\{1, 2, 3, 4, 5, 6\}$. When removing the node $4$ the graph can be covered by the sets of connected nodes: $\sigma_1=\{1,2,3\}$ and $\sigma_2=\{5, 6\}$. Coalition decomposition (Equation \ref{eq:grapheq}) allows to compute $v\left(\sigma_1 \cup \sigma_2\right) = v(\sigma_1) + v(\sigma_2)$.}
\label{fig:cgraphc}
\end{figure}

For instance in Figure \ref{fig:cgraphc} $v\left(K = \left(\sigma_1 \cup \sigma_2\right)\right)=v(\sigma_1)+v(\sigma_2)$. This decomposition would not be possible if the game was not graph-constrained since in that case, %
even when imagining a graph structure, there would be edges joining coalition $\sigma_1$ and coalition $\sigma_2$.

\begin{definition}[Myerson value \cite{myerson1977graphs, myerson1980conference}]\label{def:mye}\textit{The Myerson value of a player $i \in \mathcal{C}$ in a transferable utility coalitional game over a graph $G_{\Gamma}$ is indeed defined as the Shapley value (Equation \ref{eq:shapley}) of the graph-constrained game.}
\end{definition} Myerson provided a first axiomatization of the problem of allocating importance to the members in a graph-constrained transferable utility coalitional game in terms of equity, efficiency, and fairness \cite{myerson1977graphs}. Later, Myerson \cite{myerson1980conference} demonstrated that the only allocation rule for importance in graph-constrained transferable utility coalitional games that abides by all the necessary properties is equivalent to the computation of the Shapley values.

Computationally, the key idea is that the property in Equation \ref{eq:grapheq} can be exploited to execute fewer computations since many subsets $K \in \mathcal{P}(\mathcal{C})$ may share the same connected components $\mu(K)$. %

It is worth noting that computing the theoretical optimal exact Shapley values is $O(2^{\lvert\mathcal{C}\rvert})$ while computing Myerson values is $O(2^{M})$ with $M \leq \lvert\mathcal{C}\rvert$ being a constant proportional to the minimum number %
of connected nodes covering the graph $\Gamma$ needed to form any coalition.
Indeed, in Algorithm \ref{algo:myerson} we can see how Dynamic Programming can exploit the graph structure performing fewer computations: once each coalition $K$ with and without a feature $i$ is decomposed into the minimum number of sets of connected nodes covering $\Gamma$ (Lines 4-5), a dictionary containing the already computed value for every small coalition can be expanded %
by calculating the values starting from coalitions with increasing size (Lines 6-15). When possible, Coalition Decomposition (Equation \ref{eq:grapheq}) is exploited (Line 10 and Line 14). %

The computation of Myerson values to explain the contribution of both attributes and policies suits the task of explainability in MASs where an a priori hierarchical structure between agent characteristics and policies can be trivially inferred just from the rules of the game.
Let us put an example to explain a MAS learning to play football. We can imagine for example that the passing precision of a midfield agent interacts with the ball control ability of the forward agent, but it may not (as strongly) interact with the passing precision of the forward. This is why a causal graph of the main a priori features that can be measured can help speed up the computation of characteristic functions to explain the model.
In the next section, we provide a protocol to build such a graph.

Not every transferable utility coalitional game benefits from a prior domain knowledge structure, such as a graph, that restricts the interaction among players.  %
However, when this is possible, Myerson %
values allow one to take advantage of the graph structure to compute each contribution beyond the participation of each player, i.e., explaining the relevance of both features and policies of agents in a multi-agent environment in terms of a hierarchy of interacting features.

\section{Hierarchical Knowledge Graph (HKG) for Multi-Agent Systems (MASs) with static, dynamic, passive and active features\label{seq:tree}}
A graph structure can be introduced to describe the interaction between agents in a MAS. This practice is not uncommon in the context of Agent-Based Modeling, where the purpose is to formally describe the interaction between agents (e.g., \cite{kurve2013agent,rai2015graph,moya2017agent,robles2021multimodal}). But our purpose is to describe the interactions not only between agents but also between their individual features. For example, an agent is endowed with a policy (a set of rules, a function, a black box that tells him how to act) and individual features. Individual features can be intuitively separated into dynamic (that change with time) and static (that are constant during a game run). Static attributes can be further divided into active, passive, and necessary attributes. Necessary attributes are usually the most important in the hierarchy and are required for a policy to be deployed: e.g., if there is a value for $\textit{Max Health Points}$ in the game, then it %
becomes evident that an agent needs this feature to assume a value $>0$ in order to act in the environment since a dead agent cannot act. Therefore the $\textit{Maximum Health Points}$ are a necessary attribute since if they %
are set to zero, a player cannot exist.
Formally, the set of Static Necessary Attributes $\mathcal{N}_j$ of an agent $j$ can be defined as follows.
\begin{definition}[Static Necessary Attributes of an Agent]
Let $A_j$ be the set of all attributes and policy relative to the $j$-th agent, then the set $\mathcal{N}_j \subseteq A_j$ of Static Necessary Attributes of the said agent is
\begin{equation}
\label{eq:necessary}
\mathcal{N}_j = \left\{i_j\hspace{0.1cm} \big\vert\hspace{0.1cm} i_j \in A_j \land \nu(K\backslash\{i_j\}) =\nu(K\backslash A_j), \forall K \subseteq \mathcal{C} \right\}
\end{equation}
where $K$ is any subset of the set $\mathcal{C}$ of players in the coalitional game.
\end{definition}
In layman's terms, evaluating the characteristic function on a coalition without any element of $\mathcal{N}_j$ is equal to evaluating the characteristic function on the same coalition without every feature relative to the agent.
Usually, it should be easy for a practitioner that intends to apply the method to identify at least a subset of features in $\mathcal{N}_j$ for some agent $j$. However, identifying the full list of features that abide by Equation \ref{eq:necessary} is not mandatory.

\textit{Static active} attributes are the ones that are expressed by the agent only through directly acting in the system: e.g., in football, the shooting accuracy of a player is expressed only when the player shoots. %
\textit{Static passive} attributes can be expressed by interacting with the environment or with other agent policies: e.g., in football, resistance to pushes can be expressed only when another agent is pushing the player. In this way, %
all that is needed is basic prior knowledge of the game rules to create an HKG that includes agent-wise partitioned features and policies.

\begin{figure}[ht]
\centering
\resizebox{\columnwidth}{!}{%
\begin{tikzpicture}
[
    level 1/.style = {red},
    level 2/.style = {blue},
    level 3/.style = {teal},
    every node/.append style = {draw, anchor = west},
    grow via three points={one child at (0.5,-0.8) and two children at (0.5,-0.8) and (0.5,-1.6)},
    edge from parent path={(\tikzparentnode\tikzparentanchor) |- (\tikzchildnode\tikzchildanchor)}]
 
\node {}
    child {node {Agent 1 Static Necessary Attributes}
    child {node {Agent 1 Policy}
    child {node {Agent 1 Static Active Attributes}}}
    child [missing] {}
    child {node {Agent 1 Static Passive Attributes}}
    edge from parent node [draw = none, left] {}}
    child [missing] {}
    child [missing] {}
    child [missing] {}
    child {node {Agent 2 Static Necessary Attributes}
    child {node {Agent 2 Policy}
    child {node {Agent 2 Static Active Attributes}}}
    child [missing] {}
    child {node {Agent 2 Static Passive Attributes}}
    edge from parent node [draw = none, left] {}};
\end{tikzpicture}%
}
\caption{Example of Hierarchical Knowledge Graph for a Two Agent System. Same levels in the hierarchy are represented by the same color. Group of attributes with the same characteristics (same agent, active/passive) are fully connected between themselves). Edges do not connect boxes of the same level but only between parents and children. The top parent box is needed to represent interactions between static necessary attributes of different agents.}
\label{fig:tree}
\end{figure}
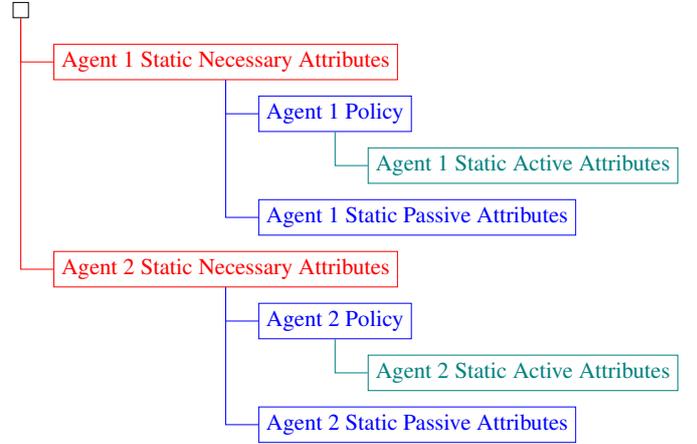

An HKG for a MAS is built as follows, having:
\begin{enumerate}
    \item Fully connected interaction between static active attributes of the same agent;%
    \item Every static active attribute of the agent is connected to the agent policy, which is always a node in the HKG;
    \item Agent's static necessary attributes are connected to the agent policy;
    \item All static necessary attributes of the game (of all agents) are fully connected between them;
    \item Passive attributes of an agent are fully connected between them;
    \item Passive attributes of each agent are connected to the necessary attributes of the agent.
\end{enumerate}
\begin{remark}It is worth noting that the HKG is hierarchical uniquely in the sense that there is a "layer-like" grouping %
per agent attributes and policies (see Figure \ref{fig:tree}). Nevertheless, since both attributes of different agents and attributes of the same agent within the same group (e.g., Static Active Attributes) interact with each other, an HKG, despite its hierarchical structure, is a graph and not a tree.\end{remark}
In Figure \ref{fig:tree} an HKG for a Two Agents System is displayed. The former can be seen as a general template for feature-policy interaction in a MAS. However, even though it is easily understandable, its validity is not general. One can imagine a system in which combinations of features interact with each other in the most complex and non-linear ways. If %
features and policies are weakly interacting, then the Myerson analysis protocol might provide a good approximation of the true interaction model. Otherwise, other means of including the prior information must be found to leverage the isolation of subgroups of features and policies that Myerson's HKG leverages to speed up the analysis.

Since the computational cost due to the number of operations needed to compute Myerson values scales with the number of connected components in the graph,%
 the more interaction between features, the more edges in the graph. This means that the game is less constrained, hence there are more connected components and thus the computational benefits of this approach (related to the ability to isolate connected components of features or sub-HKG) are restrained. 
\begin{remark}According to the protocol we put every feature in the graph. What's important is where. If a practitioner commits a mistake and erroneously puts any feature above its "true" position in the hierarchy, then the Myerson method will be compromised (i.e. considering a non-necessary attribute as a necessary one). Nevertheless, the approach is stable with respect to the opposite mistake: if a practitioner puts a feature below its true position in the hierarchy, the approach will still yield correct results.
Placing a feature up in the hierarchy means considering that all the features below it in the HKG are determined by its presence. When the said feature is not in a coalition, then the coalition will be evaluated as 0.
Placing a feature down in the hierarchy implies that fewer attributes depend hierarchically on it. The method will not be able to exploit the full potential hierarchy to perform fewer computations, but the results will be valid.
\end{remark}
\section{Requirements to compute Shapley and Myerson values for Multi-Agent Systems using a Simulator}
\label{seq:sim}
We will outline the requirements to compute the Shapley and Myerson values in a MAS using a simulator for rollouts.

\textbf{Replacement rule: }\begin{algorithm}
\caption{Exact Shapley Values Roll-out Computation with Simulator $f$}\label{algo:shapley-sim}
\KwInput{$f$ simulator of the characteristic function $v$ of the coalitional game $(\mathcal{C},v)$, $N$ number of simulations, $\zeta$ replacement rule}
\KwOutput{$n=1$ \KwTo $N$, $i=1$ \KwTo $\lvert\mathcal{C}\rvert$ values $\phi^{n}_i(v)$ Shapley values}
 \textbf{Initialization:} $\phi^{n}_{i}(v)= 0$ with $n=1$ \KwTo $N$\\
 \For{$i \in \mathcal{C}$}{
 Generate the power set $\mathcal{P}\left(\mathcal{C}\backslash\{i\}\right)$\\
 \For{$K \in \mathcal{P}\left(\mathcal{C}\backslash\{i\}\right)$}{
  \For{$n=1$ \KwTo $N$}{
 $\phi^{n}_i(v) \leftarrow \phi^{n}_i(v) + \frac{\lvert K\rvert! \left(\lvert \mathcal{C}\rvert - \lvert K\rvert -1\right)!}{\lvert \mathcal{C}\rvert!} \left(f\left(\zeta(K\cup \{i\})\right) - f\left(\zeta(K)\right)\right)$ }
 }}
\end{algorithm}
In order to compute both Shapley and Myerson values using a simulator it is mandatory to define a replacement rule for features and policies.
Indeed, a simulator $f$ of the MAS will likely require that all the agents are well-defined: they possess the full list of attributes. This means that $f:\{\sigma\text{ s.t. } \lvert\sigma\rvert = \lvert\mathcal{C}\rvert\}\rightarrow \mathbb{R}$. The simulator can only evaluate coalitions with the same number of players as the full number of features and policies to be analyzed.

How to deal with this hindrance? What is important is that the transferable utility coalitional game respects the properties of Definition \ref{def:coal} and hence that Equation \ref{eq:null} is valid.

We should then find a set of valid features/policies for the MAS $\Xi = \{\xi_1, \dots, \xi_{\lvert\mathcal{C}\rvert}\}$ with $\lvert\Xi\rvert = \lvert\mathcal{C}\rvert$ such that $f(\Xi) = 0$. Notice that $\xi_i$ is not necessarily an element of $\mathcal{C}$, what is important is that $f$ is well-defined when the feature/policy $i$ has value $\xi_i$.

In this way we can imagine that $\Xi$ is equivalent to the void coalition $\varnothing$ with respect to the characteristic function $v$: $\Xi \stackrel{v}{\leftrightarrow} \varnothing$.

We will put into a 1-to-1 relationship any $\sigma \in \mathcal{P}(\mathcal{C})$ such that $\lvert \sigma \rvert < \lvert \mathcal{C}\rvert$ and the coalition:
\begin{equation}
\label{eq:replacement-rule}
    \zeta(\sigma) = \sigma \cup \{\xi_i \text{ }| i \in \mathcal{C} \land i \not\in \sigma\}.
\end{equation}
Notice that $\lvert \zeta(\sigma) \rvert = \lvert \mathcal{C} \rvert$ and $f(\zeta(\sigma))$ is well-defined.
We will assume that $\sigma \stackrel{v}{\leftrightarrow} \zeta(\sigma)$.

Algorithms \ref{algo:shapley} and \ref{algo:myerson} can be adapted to take care of the simulator $f$ and the replacement rule $\zeta$. 
In particular Algorithm \ref{algo:shapley-sim} (and \ref{algo:myerson-sim}) performs Shapley (and Myerson) analysis using a simulator $f$ for a number of roll-outs (simulations) $N$ given as input. Notice that the purpose of the simulator is providing the value of the characteristic function $\nu$, otherwise not computable. Additionally, as stated before, the simulator can only work if all agents are well-defined (with a valid policy and a full list of attributes). In order to evaluate the characteristic function at coalitions without a subset of policies and/or attributes, line 6 is called in Algorithm \ref{algo:shapley-sim}, and lines 10, 14, and 16 are called in Algorithm \ref{algo:myerson-sim}. These lines perform the composition of functions $f \circ \zeta$: first, $\zeta$ replaces accordingly the removed features/policies in order to deal, successively, with a simulation $f$ applied on a well-defined environment. Several simulation roll-outs $N$ will be performed to explore the distribution of the results generated by the stochasticity of the environment.

\begin{remark}
Note that the results of this analysis provide intuition about the contribution of policies and attributes. However, even if an attribute is deemed more important than another, it is likely, but not granted, that the very same MAS where that single attribute is assuming a greater Shapley/Myerson value will be more performing. This is why this approach should not be used online during the learning phase of RL agents but only a posteriori, since credit assignment will only be fully given at the end of the task/game.
\end{remark}

\begin{algorithm}
\caption{Exact Myerson Values Roll-out computation with Simulator $f$\label{algo:myerson-sim}}
\KwInput{$\Gamma$ graph over the set of players $\mathcal{C}$ of the coalitional game $(\mathcal{C},v)$, $f$ simulator of the characteristic function $v$ of the coalitional game $(\mathcal{C},v)$, $N$ number of simulations, $\zeta$ replacement rule}
\KwOutput{$n=1$ \KwTo $N$, $i=1$ \KwTo $\lvert\mathcal{C}\rvert$ values $\phi^{n}_i(v)$ Myerson values}
 \textbf{Initialization:} $\phi^{n}_{i}(v)= 0$ with $n=1$ \KwTo $N$, Coalitions = $\{\varnothing\}$\\
 \For{$i \in \mathcal{C}$}{
 Generate the power set $\mathcal{P}\left(\mathcal{C}\backslash\{i\}\right)$\\
 Decompose each $K \in \mathcal{P}\left(\mathcal{C}\backslash\{i\}\right)$ to $\mu(K)$, the sets of connected nodes minimally covering the subgraph with vertices $K$\\
 Decompose each $K \cup \{i\}$ to $\mu(K \cup \{i\})$, the sets of connected nodes minimally covering the subgraph with vertices $K \cup \{i\}$\\
 \For{$K \in \mathcal{P}\left(\mathcal{C}\backslash\{i\}\right)$ ordered by increasing $\lvert K \rvert$}{
  \For{$n=1$ \KwTo $N$}{
  \If{$\zeta(K) \not\in$ Coalitions}
  {\For{$\sigma \in \mu(K) \land \zeta(\sigma) \not\in\text{Coalitions}$}{Coalitions$(\zeta(\sigma)) \leftarrow f(\zeta(\sigma))$}Coalitions$(\zeta(K)) \leftarrow \sum_{\sigma \in \mu(K)}\text{Coalitions}(\zeta(\sigma))$}
\If{$\zeta(K \cup \{i\}) \not\in$ Coalitions}
  {\For{$\sigma \in \mu(K\cup\{i\}) \land \zeta(\sigma) \not\in\text{Coalitions}$}{Coalitions$(\zeta(\sigma)) \leftarrow f(\zeta(\sigma))$}Coalitions$(\zeta(K\cup\{i\})) \leftarrow \sum_{\sigma \in \mu(K\cup\{i\})}\text{Coalitions}(\zeta(\sigma))$}
 $\phi^{n}_i(v) \leftarrow \phi^{n}_i(v) + \frac{\lvert K\rvert! \left(\lvert \mathcal{C}\rvert - \lvert K\rvert -1\right)!}{\lvert \mathcal{C}\rvert!}$ \resizebox{.35\textwidth}{!}{%
 $\left(\text{Coalitions}\left(\zeta(K\cup \{i\})\right) - \text{Coalitions}\left(\zeta(K)\right)\right)$%
 }}
 }}
\end{algorithm}

\section{Experimental Contribution}
\label{seq:exp}
In Section \ref{seq:def} we defined a transferable utility coalitional game between players. From now on we will use the word player to refer to actual members of a team in our game that cooperate to reach the common goal. We will define the transferable utility coalitional game over a set of players $\mathcal{C}$ made of features (individual attributes and policies).

We propose to exploit prior knowledge about the transferable utility coalitional game to be analyzed, in order to build up a connected graph in which each node represents an agent's individual attribute or policy.
This approach extends \cite{heuillet2022collective} since 1) in that work only agents' policies were analyzed and directly included in the paradigm; and 2) we introduce the HKG as domain knowledge to be exploited. 

We test our approach in a game setting (that is richer %
than the Particle environment used in \cite{heuillet2022collective} since the agents possess several individual static attributes), \textit{Arena}\footnote{The program code is open and available in the GitHub repository: \url{https://github.com/giorgioangel/myersoncoop}.}, a game inspired by World of Warcraft 3 vs 3 arena matches. In order to compute Shapley and Myerson values we will perform rollout simulations using a simulator of the game. Several simulations are needed due to the stochasticity of the environment and sometimes of the policies. The statistical significance of the results will be validated by the Mann-Whitney U test which is a non-parametric statistical test specifically suited to compare the equality, in probability, of two populations. %

\subsection{\textit{Arena} Game: RL Environment Description}
Two teams, team \textit{A} and team \textit{B}, both of them made of a \textit{Warrior}, a \textit{Mage}, and a \textit{Priest} fight each other. The common goal of a team is to defeat every enemy opponent in the least possible amount of moves. Given the different possible actions available to each agent, selfish strategies may easily lead to defeat. %
The teams perform their sequence of actions taking turns one at a time.
At the beginning of each match, one team is chosen to start first with a random uniform probability. The agents in each team act abiding by the following order: 1) Warrior, 2) Mage, 3) Priest. %

\textbf{Victory Condition}: The Arena game ends when all agents in a team are dead or when $T=1000$ rounds have passed.
Let $\Omega$ be the set of all possible simulations. When team \textit{A} wins, a simulation $\omega \in \Omega$ the game returns a result of $r(\omega)=-1$, when team \textit{B} prevails the returned result amounts to $r(\omega)=1$, if $T(\omega)=1000$ rounds have passed and there is still not a winner then $r(\omega)=0$. 

The final score of a simulation is: \begin{equation}
\label{eq:score}
    S(\omega) = 100\left(\dfrac{r(\omega)}{T(\omega)}+1\right)
\end{equation}
where $T(\omega)$ is the total number of rounds needed to terminate the game. Notice that $S \in [0, 200]$. If team \textit{A} wins in one round, $S=200$, if it loses in one round, then $S=0$. Furthermore, $S\rightarrow 100$ with $T\rightarrow 1000$. Intuitively,  we will have $200\geq S>100$ for simulations where team \textit{A} won and $0\leq S <100$ where it lost. %

\textbf{Description of agents' roles and attributes}: Every agent executes a \textit{policy}. We assume that the full list of individual attributes is known. We build an HKG for such a MAS with the individual attributes and policies following the protocol described in Section \ref{seq:tree}. each agent's attributes are divided into Static Necessary Attributes, Static Active Attributes, and Dynamical Attributes:
\begin{enumerate}
    \item Max Health Points [Static Necessary Attribute]: the maximum health points that an agent can possess;
    \item Attack Power [Static Active Attribute]: the maximum damage that an agent can deal within one time step;
    \item Healing Power [Static Active Attribute]: the maximum amount of health points that an agent can lend by healing himself or another one in one time step;
    \item Control Chance [Static Active Attribute]: modulates the chance the \textit{Mage} has to stop other agents from acting from one round (will be defined better later in Equation \ref{eq:magecontrol}); %
    \item Current Health Points [Dynamical Attribute]: the health points of an agent at each time step.
\end{enumerate}

The default value of these individual attributes is reported in Table \ref{tab:default_stats}. We do not consider Static Passive Attributes. Subsequently to this work, we tried to add to the game a Defense attribute for each agent. Results were promising but the increase of features in the transferable utility coalitional game from $15$ to $18$, while still manageable for the Myerson method with HKG, made the problem already computationally too demanding for the Shapley approach in terms of our available computational resources.
\begin{table}[htbp!]
\caption{Default individual attributes (static and variable during the game) of each agent in the \textit{Arena} cooperative multi-agent environment.}
\label{tab:default_stats}
\centering
\begin{tabular}{lc}
\textbf{Static Necessary Attribute}       & \textit{\textbf{Value} [Range]} \\ \hline
& \\
\textit{MaxHealthPoints} & 100 [0-100]                      \\
 & \\
\textbf{Static Active Attributes}       & \textit{\textbf{Value} [Range]} \\ \hline & \\
\textit{AttackPower}     & 10 [0-20]                              \\
\textit{HealingPower}    & 5 [0-100]                              \\
\textit{ControlChance}   & 0.5  [0-0.5]                          \\ & \\
\textbf{Dynamical Attribute}       & \textit{\textbf{Value} [Range]} \\ \hline & \\
\textit{CurrentHealthPoints} & 100 [0-\textit{MaxHealthPoints}]          
\end{tabular}
\end{table}

\subsubsection{Warrior}
The warrior can only \textit{attack} an enemy agent.
He damages the enemy by an amount equal to his (the warrior's) \textit{AttackPower}:
\begin{equation}
\label{eq:wardamage}
    \textit{tarHP}_{t+1}=\max\left(0,\textit{tarHP}_{t}-\textit{AttackPower}\right)
\end{equation}
where \textit{tarHP} represents the \textit{CurrentHealthPoints} of the targeted enemy and $t$ a time step.
Any agent dies when his \\$\textit{CurrentHealthPoints} = 0$.
\subsubsection{Mage}
A mage can only \textit{control} (put to sleep) an enemy agent.
His chance $P$ of controlling the enemy is equal to
\begin{equation}
\label{eq:magecontrol}
   P = \textit{ControlChance} \left(1 + \frac{\textit{AttackPower}}{20}\right).
\end{equation}
When an enemy agent is put to sleep he cannot perform any action during the next turn.
\subsubsection{Priest}
A priest can only \textit{heal} a teammate. He heals the teammate by paying an amount equal to his %
\textit{HealingPower}:
\begin{equation}
\label{eq:priestheal}
    \textit{tarHP}_{t+1}=min\left(\textit{tarHP}_{t}+\textit{HealingPower},\textit{tarMaxHP}\right)
\end{equation}
where \textit{tarMaxHP} are the \textit{MaxHealthPoints} of the targeted agent and \textit{HealingPower} is the one of the \textit{Priest}.%
\subsection{Policies}
Three different handcrafted policy types are enabled for all agents both in the hardcoded policy setting and, later on, in the setting with a policy learned with RL:
\begin{enumerate}
    \item Random: with this policy, the target of the warrior and the mage are uniformly chosen between the alive enemies.
The target of the priest is uniformly chosen between the alive teammates. 
    \item Smart: the Warrior and Mage target the living enemies with the following priority list: 1) Priest, 2) Mage, 3) Warrior. %
The Priest always heals the living teammate with the least \textit{CurrentHealthPoints}.%
    \item Do nothing (\textit{No-Op}): While following this strategy, the agent does not perform any action.
    \item Deep Reinforcement Learning (\textit{RL}): Intending to show that the proposed approach can provide reasonable explanations of black-box Reinforcement Learning models based on Deep Neural Networks we trained a Stable-Baselines3's A2C model \cite{stable-baselines3} where every agent can select the target of his action at every time step. Therefore the \textit{Warrior} will decide who to hit between the enemies, the \textit{Mage} which enemy to control, and the \textit{Priest} who to heal among his friends. Immediately then each agent will choose which action to perform on it. We trained the model in three phases, every time until convergence. In Phase 1 the enemy team was %
hardcoded to deploy a \textit{No-op} policy, in Phase 2 the enemy team %
hardcoded to act following the \textit{Random} policy and in Phase 3 the enemy team was hard-coded to follow a \textit{Smart} policy. In this way, we could provide the A2C agent an adversary with increasing difficulty along the three phases. Moreover, the reward signal used to train the agent was not the sparse final score of Equation \ref{eq:score} but the difference between the total current health points of the teams. The said tricks let the training converge faster. 

With little surprise, the policy learned by the A2C model managed to overpower every hand-crafted policy with a 100\% victory rate.

We noticed that the A2C acts in the following way: the A2C trained agent learns to control all three agents in its team. The \textit{A2C Warrior} and the \textit{A2C Mage} both learn to attack and control the enemy's \textit{Warrior}. Whenever this last one dies, the \textit{A2C Mage} controls the enemy's \textit{Priest} while the \textit{A2C Warrior} attacks indiscriminately one between the remaining living enemies. During the whole match, the \textit{A2C Priest} heals whoever of his team is taking damage.
\end{enumerate}

\subsection{Research Question Hypothesis}

Hypothesis: we want to show that it is possible to explain both the importance of individual policies and the individual static attributes of agents in a MAS. With this in mind, we first build a transferable utility coalitional game whose players are both the policies and the individual features (separately taken), then we constrain the game onto an HKG: a graph structure for the MAS that is built following the protocol provided in Section \ref{seq:tree}.
In order to test whether the approach is valid, we will compute both the Shapley Values (without the knowledge graph) and the Myerson values (exploiting the knowledge graph). In both cases, the characteristic function will be given by the score $S$ which is the output of a game simulator $f$ (see Equation \ref{eq:score}). Since the simulator requires every agent to be well defined (with a valid policy and a full set of attributes), we have to apply the replacement rule $\zeta$ to coalitions before running it (see Equation \ref{eq:replacement-rule}). Hence, after applying $\zeta$, every coalition will be legitimate in the sense that applying $f$ (the simulator) to them will produce a result (a real number). However, in Shapley analysis, we will first use the replacement rule and then run the simulation (line 6 of Algorithm \ref{algo:shapley-sim}), while in Myerson analysis we will first check for the graph connectivity in order to exploit Property 5 in Definition \ref{def:mye} (lines 4 and 5 in Algorithm \ref{algo:myerson-sim} decompose the coalition subgraph in connected parts and then the simulation is run only for the decomposed coalitions).
We will show that:
\begin{enumerate}
\item by defining different yet correct replacement rules for policies and attributes the two can stand on the same footing with respect to these analyses;
    \item the $N=72$ results of Shapley and Myerson analysis come from the same distribution, and therefore that the HKG provides a good approximation to a latent structure of a MAS;
    \item both the Shapley and the Myerson values are consistent with the rules of the game and the predicted contribution seems reasonable;
    \item the number of computations needed to carry out Myerson analysis is lower than the Shapley one. Indeed computing exact Shapley values is $O(2^{\lvert\mathcal{C}\rvert})$ while computing Myerson values is $O(2^{M})$ with $\mathcal{C}$ being defined in Definition \ref{def:coal} as the number of players and $M \leq \lvert\mathcal{C}\rvert$  being a constant proportional to the minimum number of connected nodes covering the graph $\Gamma$ needed to form any coalition. In our particular case, $\lvert\mathcal{C}\rvert = 15$, $2^{15}=32768$ while $M \approx 9.966$ and $2^{9.666} \approx 1000$.

\end{enumerate}
All players share the same individual stats as reported in Table \ref{tab:default_stats}.

Let us extend the formalization of the game score $S$ defined in Equation \ref{eq:score}. Let $\mathcal{P}\left(\mathcal{C}\right)$ be the power set of $\mathcal{C}$, the set of static attributes (Table \ref{tab:default_stats})
 and policies of team \textit{A}. We define $\Omega_{\sigma}$ with $\sigma \in \mathcal{P}\left(\mathcal{C}\right)$ as the set of possible simulations for a specific coalition $\sigma$. When an individual attribute is not present in $\sigma$, then it is set to zero before starting the simulation (e.g. a coalition without \textit{Warrior's AttackPower} means that in simulation the warrior will start the simulation with $\textit{AttackPower} = 0$). When a policy is not present, it is set to \textit{Do nothing}. 

Notice that, if the policy and the features of \textit{team B's Warrior} allow him to deal damage, %
then the score of every possible simulation $\omega_{\emptyset} \in \Omega_{\emptyset}$ where $\Omega_{\emptyset}$ is the set of simulations' outcomes attainable with an empty coalition is %
$S(\omega_{\emptyset}) = 0$%
. Let us define an average score function over a set of $N$ simulations $\Sigma : \mathbb{N}_+, \times \mathcal{P}\left(\mathcal{C}\right) \rightarrow [0,200]$,
\begin{equation}
\label{eq:simscore}
    \Sigma(n, \sigma) = \dfrac{1}{n}\sum_{i=1}^{n}S(\omega_{\sigma, i}).
\end{equation}
where $\omega_{\sigma,i}$ is the outcome of the $i$-th simulation run with coalition $\sigma$.

Our goal is to compute the importance of the individual \textit{static} attributes and policies of each member in team \textit{A}.
We will run $N=72$ simulations with \textit{team A } and \textit{team B} playing all the possible combinations of policies in the set: \newline$\{\text{Random, Smart, No-Op, RL}\}$.

We will compute these values with two different approaches: 1) naively calculating the Shapley values, %
2) computing the Myerson values on a properly crafted HKG (Figure \ref{fig:game-tree}).

\subsection{Replacement rule}
\label{ss:repl}
In the case of the \textit{Arena} game \\$\xi_{policy} = \textit{No-Op}$ $\forall \textit{policy} \in \{\textit{Smart, Random, No-Op, RL}\}$ and $\xi_{j} = 0$ $\forall j \in \mathcal{C} \land j \not\in \{\textit{Smart, Random, No-Op, RL}\}$, or in other words when $j$ is an attribute and not a policy.

\subsection{Shapley values}
In order to compute the Shapley values we use as \textit{characteristic function} $v$ the score $S$ (Equation \ref{eq:score}). We use a simulator of the game $f$ and perform $N=72$ different simulations. The values are computed using Algorithm \ref{algo:shapley-sim} and the replacement rule defined in Subsection \ref{ss:repl}.

The game's mean score $\Sigma$ over $N=72$ simulations is displayed in the results Table \ref{tab:results}.
Notice that this characteristic function abides by all the properties defined in Section \ref{seq:def}.
In the first part of Table \ref{tab:results}, we report the Shapley and the Myerson values obtained for all the features and policies when \textit{team A} is playing a \textit{Smart} policy along with their computational times, in the second part of Table \ref{tab:results} \textit{team A} is acting following the \textit{Random} policy, in the third part of Table \ref{tab:results} the \textit{No-Op} policy and the last part of Table \ref{tab:results} the \textit{RL} agent. %

\subsection{Myerson values}
\begin{figure}[ht]
\centering
\resizebox{\columnwidth}{!}{%
\begin{tikzpicture}
[
    level 1/.style = {red},
    level 2/.style = {blue},
    level 3/.style = {teal},
    every node/.append style = {draw, anchor = west},
    grow via three points={one child at (0.5,-0.8) and two children at (0.5,-0.8) and (0.5,-1.6)},
    edge from parent path={(\tikzparentnode\tikzparentanchor) |- (\tikzchildnode\tikzchildanchor)}]
 
\node {}
    child {node {Warrior Max Health Points}
    child {node {Warrior Policy}
    child {node (WAP) {Warrior Attack Power}}
    child {node (WHP) {Warrior Healing Power}}
    child {node (WCC) {Warrior Control Chance}}}
    child [missing] {}
    child [missing] {}
    edge from parent node [draw = none, left] {}}
    child [missing] {}
    child [missing] {}
    child [missing] {}
    child [missing] {}
    child {node {Mage Max Health Points}
    child {node {Mage Policy}
    child {node (MAP) {Mage Attack Power}}
    child {node (MHP) {Mage Healing Power}}
    child {node (MCC) {Mage Control Chance}}}
    edge from parent node [draw = none, left] {}}
    child [missing] {}
    child [missing] {}
    child [missing] {}
    child [missing] {}
    child {node {Priest Max Health Points}
    child {node {Priest Policy}
    child {node (PAP) {Priest Attack Power}}
    child {node (PHP) {Priest Healing Power}}
    child {node (PCC) {Priest Control Chance}}}
    edge from parent node [draw = none, left] {}};
    \path[-,draw,thick, teal]
    (PAP) edge (PHP)
    (PHP) edge (PCC)
    (MAP) edge (MHP)
    (MHP) edge (MCC)
    (WAP) edge (WHP)
    (WHP) edge (WCC)
    (PAP.east) edge [bend left] (PCC.east)
    (MAP.east) edge [bend left] (MCC.east)
    (WAP.east) edge [bend left] (WCC.east);
\end{tikzpicture}%
}
\caption{Hierarchical Knowledge Graph for the \textit{Arena} game. Notice how the subgraph of Static Active Attributes is fully connected. In this game there are no Static Passive Attributes.}
\label{fig:game-tree}
\end{figure}
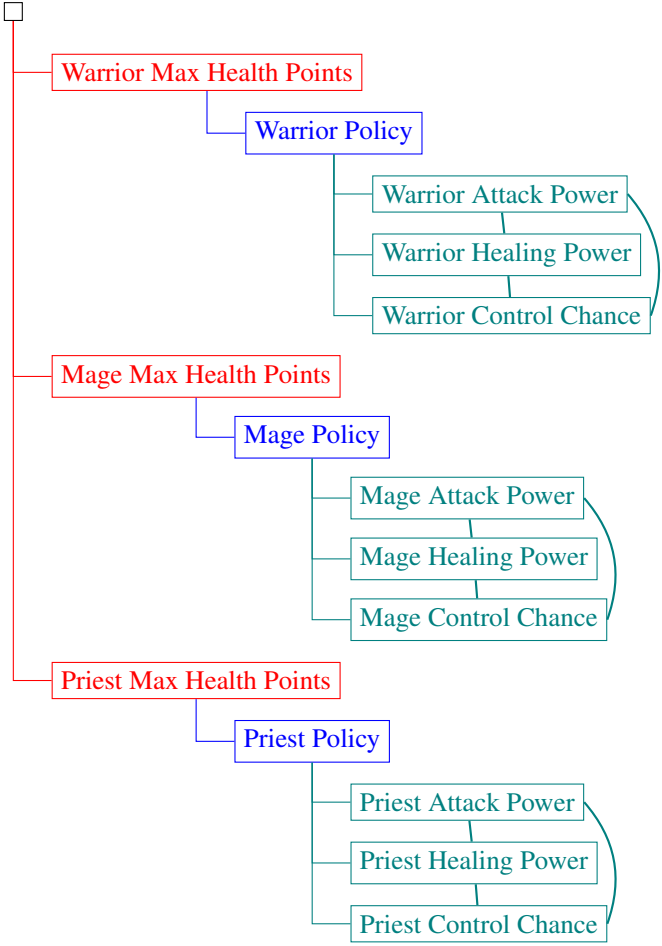

In order to compute the Myerson values we first have to define a graph $\Gamma$ that encompasses the relationship between the features. Using our prior knowledge about the game: if an agent has $\textit{MaxHealthPoints}=0$ he is already dead and then he is unable to act. If a policy is \textit{Do nothing} then all the other individual attributes besides \textit{MaxHealthPoints} do not matter, and thus, by following the protocol provided in Section \ref{seq:tree} we can build $\Gamma$ as the HKG shown in Figure \ref{fig:game-tree}. 

It is important to remember that the characteristic function $v$ of a coalition $\sigma$ of features or policies defined over the graph is the sum of the characteristic functions of the connected components of $\sigma$ (Algorithm \ref{algo:myerson}).%

Hence, if for example a coalition $\sigma$ is the whole $\mathcal{C}$ without the \textit{Warrior's MaxHealthPoints}, we will have two connected components: a coalition $\sigma_1$ with all the attributes and policies of the \textit{Mage} and the \textit{Priest}, and $\sigma_2$, a coalition with just the policy and attributes of the \textit{Warrior}. When performing a rollout for $\sigma_2$ all \textit{MaxHealthPoints} are put to zero, hence all agents are dead, and $S$ is trivially zero.

More generally, this is due to the structure of the HKG which allows to completely ignore an agent, its policy, and its attributes when the high-level node in the graph hierarchy is not part of the considered coalition.

It is worth remembering that to compute the Shapley or the Myerson values we have to consider coalitions without some participants. This is not feasible when using a simulator. Therefore we had to define a rule $\zeta$ to replace attributes and policies with something else that for us is equivalent to a coalition without that given element. This rule was replacing an attribute value with zero and the policy with the \textit{No-Op}.

If the \textit{MaxHealthPoints} feature is present, but not the policy (i.e., the policy replaced by \textit{No-Op}), and any subset of other attributes. %
This greatly reduces the number of computations to be performed in order to obtain the Myerson values that were computed using Algorithm \ref{algo:myerson-sim}.
As for Shapley analysis, the results are reported in Table \ref{tab:results}.
The number of non-trivial evaluations of the characteristic function needed to assess the Shapley values in this example is $32768$ while computing it for the Myerson values happens only $1000$ times. It is worth noting that these values are specific to this scenario and environment, and the degree of reduction could change for a different environment that would be differently described by a different HKG.

\begin{table}[H]
\caption{\tiny Rollout computation of Shapley and Myerson values to explain the contribution of each static attribute and policy. Team A is playing the \textit{Smart}, the \textit{Random} and the {No-op} policy. The time elapsed to compute the whole set of values for $N=72$ simulations is reported below the label of each column. The displayed results are averaged over $N=72$ different simulations. Statistical significance of a Mann-Whitney U test to check if the distribution of results is the same in probability as the one of a null contribution is reported after the mean value (when stars are present the distributions are different): ${}^{*}$ for $p < 0.05$, ${}^{**}$ for $p < 0.01$, ${}^{***}$ for $p < 0.001$. Statistical significance of a Mann-Whitney U test between the distribution of Shapley values and the one of Myerson values is displayed in \textbf{bold} when the distributions are different with $p < 0.05$.} %
\label{tab:results} \centering
\vspace{-0.25cm}
\resizebox{0.94\columnwidth}{!}{%
    \begin{tabular}{lllllllll}
    &\multicolumn{2}{c}{Random vs Random}&\multicolumn{2}{c}{Random vs Smart}&\multicolumn{2}{c}{Random vs No-Op}&\multicolumn{2}{c}{Random vs RL}\\
    \textbf{Feature}         & \textit{\textbf{Shapley}} & \textit{\textbf{Myerson}}& \textit{\textbf{Shapley}} & \textit{\textbf{Myerson}}& \textit{\textbf{Shapley}} & \textit{\textbf{Myerson}}& \textit{\textbf{Shapley}} & \textit{\textbf{Myerson}} \\
    \textbf{Total Score $\Sigma$} & 99.97&99.87&97.18&97.17&103.33&103.33&98.26&98.34 \\\textbf{Comp. Time (s)} & 16796.07&392.50&34033.64&531.05&22642.04&760.45&40962.12&994.56 \\\hline
    &                           & & & & &                          & &\\
    \textbf{Agent: Warrior }        &                           &                           & &\\
    \textit{MaxHealthPoints} & 32.30${}^{***}$                   & 32.32${}^{***}$ & 28.25${}^{***}$ &  28.25${}^{***}$ & 34.44${}^{***}$  & 34.44${}^{***}$ & 36.79${}^{***}$  & 36.81${}^{***}$                \\
    \textit{Policy}     & 0.20${}^{***}$                    & 0.19${}^{***}$      & 0.09${}^{***}$ &   0.09${}^{***}$ & 1.11${}^{***}$ & 1.11${}^{***}$ & -0.21${}^{***}$  & -0.21${}^{***}$         \\
    \textit{AttackPower}     & 0.18${}^{***}$                    & 0.16${}^{***}$      & 0.09${}^{***}$ &   0.09${}^{***}$ & 1.11${}^{***}$ & 1.11${}^{***}$ & -0.22${}^{***}$  & -0.20${}^{***}$         \\
    \textit{HealingPower}    & -0.02                   & -0.01     & 0.00  & 0.00 & 0.00 & 0.00 & 0.00  & -0.01        \\
    \textit{ControlChance}   & 0.02                   & -0.01    & 0.00    & 0.00 & 0.00   & 0.00 & 0.00  & 0.00        \\
    &                           & & & & &                          & &\\
    \textbf{Agent: Mage }        &                           &                           & &\\
    \textit{MaxHealthPoints} & 32.55${}^{***}$                   & 32.56${}^{***}$ & 32.81${}^{***}$ &  32.80${}^{***}$ & 33.33${}^{***}$  & 33.33${}^{***}$ & 30.49${}^{***}$  & 30.52${}^{***}$                \\
    \textit{Policy}     & 0.37${}^{***}$                    & 0.36${}^{***}$      & 0.09${}^{***}$ &   0.09${}^{***}$ & 0.00 & 0.00 & 0.30${}^{***}$  & 0.30${}^{***}$         \\
    \textit{AttackPower}     & 0.19${}^{***}$                    & 0.17${}^{***}$      & 0.03${}^{***}$ &   0.03${}^{***}$ & 0.00 & 0.00 & 0.09${}^{***}$  & 0.09${}^{***}$         \\
    \textit{HealingPower}    & 0.00                   & -0.03     & 0.00  & 0.00 & 0.00 & 0.00 & 0.00  & 0.00        \\
    \textit{ControlChance}   & 0.37${}^{***}$                   & 0.35${}^{***}$    & 0.09${}^{***}$    & 0.09${}^{***}$ & 0.00   & 0.00 & 0.29${}^{***}$  & 0.28${}^{***}$        \\
    &                           & & & & &                          & &\\
    \textbf{Agent: Priest }        &                           &                           & &\\
    \textit{MaxHealthPoints} & 32.76${}^{***}$                   & 32.78${}^{***}$ & 35.70${}^{***}$ &  35.69${}^{***}$ & 33.33${}^{***}$  & 33.33${}^{***}$ & 30.20${}^{***}$  & 30.24${}^{***}$                \\
    \textit{Policy}     & 0.54${}^{***}$                    & 0.56${}^{***}$      & 0.02${}^{***}$ &   0.03${}^{***}$ & 0.00 & 0.00 & 0.26${}^{***}$  & 0.28${}^{***}$         \\
    \textit{AttackPower}     & -0.01                    & -0.04      & 0.00 &   -0.01 & 0.00 & 0.00 & 0.00${}^{*}$  & 0.01         \\
    \textit{HealingPower}    & 0.55${}^{***}$                   & 0.53${}^{***}$     & 0.02${}^{***}$  & 0.02${}^{***}$ & 0.00 & 0.00 & 0.25${}^{***}$  & 0.24${}^{***}$        \\
    \textit{ControlChance}   & 0.00                   & -0.01    & 0.00    & 0.00 & 0.00   & 0.00 & \textbf{0.01}  & \textbf{-0.02}        \\
    &                           & & & & &                          & &\\
    &                           & & & & &                          & &\\
    &\multicolumn{2}{c}{Smart vs Random}&\multicolumn{2}{c}{Smart vs Smart}&\multicolumn{2}{c}{Smart vs No-Op}&\multicolumn{2}{c}{Smart vs RL}\\
    \textbf{Feature}         & \textit{\textbf{Shapley}} & \textit{\textbf{Myerson}}& \textit{\textbf{Shapley}} & \textit{\textbf{Myerson}}& \textit{\textbf{Shapley}} & \textit{\textbf{Myerson}}& \textit{\textbf{Shapley}} & \textit{\textbf{Myerson}} \\
    \textbf{Total Score $\Sigma$} & 102.83&102.76&99.87&99.91&103.33&103.33&97.61&97.64 \\\textbf{Comp. Time (s)} & 16284.31&385.34&34129.29&534.81&24156.22&868.38&39698.93&719.89 \\\hline
    &                           & & & & &                          & &\\
    \textbf{Agent: Warrior }        &                           &                           & &\\
    \textit{MaxHealthPoints} & 32.89${}^{***}$                   & 32.90${}^{***}$ & 28.53${}^{***}$ &  28.53${}^{***}$ & 34.44${}^{***}$  & 34.44${}^{***}$ & 37.05${}^{***}$  & 37.05${}^{***}$                \\
    \textit{Policy}     & 0.73${}^{***}$                    & 0.72${}^{***}$      & 0.49${}^{***}$ &   0.43${}^{***}$ & 1.11${}^{***}$ & 1.11${}^{***}$ & -0.18${}^{***}$  & -0.17${}^{***}$         \\
    \textit{AttackPower}     & 0.73${}^{***}$                    & 0.73${}^{***}$      & 0.46${}^{***}$ &   0.41${}^{***}$ & 1.11${}^{***}$ & 1.11${}^{***}$ & -0.18${}^{***}$  & -0.17${}^{***}$         \\
    \textit{HealingPower}    & 0.00                   & 0.00     & -0.01  & 0.04${}^{*}$ & 0.00 & 0.00 & 0.00  & 0.00        \\
    \textit{ControlChance}   & -0.01                   & 0.00    & 0.04${}^{**}$    & -0.02 & 0.00   & 0.00 & 0.00${}^{*}$  & 0.00        \\
    &                           & & & & &                          & &\\
    \textbf{Agent: Mage }        &                           &                           & &\\
    \textit{MaxHealthPoints} & 32.90${}^{***}$                   & 32.91${}^{***}$ & 32.90${}^{***}$ &  32.88${}^{***}$ & 33.33${}^{***}$  & 33.33${}^{***}$ & 29.99${}^{***}$  & 30.01${}^{***}$                \\
    \textit{Policy}     & 0.13${}^{***}$                    & 0.12${}^{***}$      & 0.34${}^{***}$ &   0.30${}^{***}$ & 0.00 & 0.00 & 0.00  & 0.00         \\
    \textit{AttackPower}     & 0.04${}^{***}$                    & 0.03${}^{***}$      & 0.19${}^{***}$ &   0.27${}^{***}$ & 0.00 & 0.00 & 0.00  & 0.00         \\
    \textit{HealingPower}    & -0.01                   & 0.00     & 0.01  & -0.01 & 0.00 & 0.00 & 0.00  & 0.00        \\
    \textit{ControlChance}   & 0.13${}^{***}$                   & 0.13${}^{***}$    & 0.29${}^{***}$    & 0.33${}^{***}$ & 0.00   & 0.00 & 0.00  & 0.00${}^{*}$        \\
    &                           & & & & &                          & &\\
    \textbf{Agent: Priest }        &                           &                           & &\\
    \textit{MaxHealthPoints} & 33.36${}^{***}$                   & 33.37${}^{***}$ & 36.05${}^{***}$ &  36.03${}^{***}$ & 33.33${}^{***}$  & 33.33${}^{***}$ & 30.30${}^{***}$  & 30.30${}^{***}$                \\
    \textit{Policy}     & 0.95${}^{***}$                    & 0.93${}^{***}$      & 0.28${}^{***}$ &   0.31${}^{***}$ & 0.00 & 0.00 & 0.31${}^{***}$  & 0.31${}^{***}$         \\
    \textit{AttackPower}     & 0.00                    & 0.00      & -0.04 &   0.06 & 0.00 & 0.00 & 0.00  & 0.00         \\
    \textit{HealingPower}    & 0.98${}^{***}$                   & 0.91${}^{***}$     & 0.33${}^{***}$  & 0.31${}^{***}$ & 0.00 & 0.00 & 0.31${}^{***}$  & 0.31${}^{***}$        \\
    \textit{ControlChance}   & 0.00                   & 0.00    & -0.01    & 0.02 & 0.00   & 0.00 & 0.00  & 0.00        \\
    &                           & & & & &                          & &\\
    &                           & & & & &                          & &\\
    &\multicolumn{2}{c}{Nothing vs Random}&\multicolumn{2}{c}{Nothing vs Smart}&\multicolumn{2}{c}{Nothing vs No-Op}&\multicolumn{2}{c}{Nothing vs RL}\\
    \textbf{Feature}         & \textit{\textbf{Shapley}} & \textit{\textbf{Myerson}}& \textit{\textbf{Shapley}} & \textit{\textbf{Myerson}}& \textit{\textbf{Shapley}} & \textit{\textbf{Myerson}}& \textit{\textbf{Shapley}} & \textit{\textbf{Myerson}} \\
    \textbf{Total Score $\Sigma$} & 96.67&96.67&96.67&96.67&100.00&100.00&96.67&96.67 \\\textbf{Comp. Time (s)} & 16350.52&364.33&33549.48&511.73&18503.63&435.27&44591.67&784.45 \\\hline
    &                           & & & & &                          & &\\
    \textbf{Agent: Warrior }        &                           &                           & &\\
    \textit{MaxHealthPoints} & 32.22${}^{***}$                   & 32.22${}^{***}$ & 28.06${}^{***}$ &  28.06${}^{***}$ & 33.33${}^{***}$  & 33.33${}^{***}$ & 37.22${}^{***}$  & 37.22${}^{***}$                \\
    \textit{Policy}     & 0.00                    & 0.00      & 0.00 &   0.00 & 0.00 & 0.00 & 0.00  & 0.00         \\
    \textit{AttackPower}     & 0.00                    & 0.00      & 0.00 &   0.00 & 0.00 & 0.00 & 0.00  & 0.00         \\
    \textit{HealingPower}    & 0.00                   & 0.00     & 0.00  & 0.00 & 0.00 & 0.00 & 0.00  & 0.00        \\
    \textit{ControlChance}   & 0.00                   & 0.00    & 0.00    & 0.00 & 0.00   & 0.00 & 0.00  & 0.00        \\
    &                           & & & & &                          & &\\
    \textbf{Agent: Mage }        &                           &                           & &\\
    \textit{MaxHealthPoints} & 32.22${}^{***}$                   & 32.22${}^{***}$ & 33.06${}^{***}$ &  33.06${}^{***}$ & 33.33${}^{***}$  & 33.33${}^{***}$ & 29.72${}^{***}$  & 29.72${}^{***}$                \\
    \textit{Policy}     & 0.00                    & 0.00      & 0.00 &   0.00 & 0.00 & 0.00 & 0.00  & 0.00         \\
    \textit{AttackPower}     & 0.00                    & 0.00      & 0.00 &   0.00 & 0.00 & 0.00 & 0.00  & 0.00         \\
    \textit{HealingPower}    & 0.00                   & 0.00     & 0.00  & 0.00 & 0.00 & 0.00 & 0.00  & 0.00        \\
    \textit{ControlChance}   & 0.00                   & 0.00    & 0.00    & 0.00 & 0.00   & 0.00 & 0.00  & 0.00        \\
    &                           & & & & &                          & &\\
    \textbf{Agent: Priest }        &                           &                           & &\\
    \textit{MaxHealthPoints} & 32.22${}^{***}$                   & 32.22${}^{***}$ & 35.56${}^{***}$ &  35.56${}^{***}$ & 33.33${}^{***}$  & 33.33${}^{***}$ & 29.72${}^{***}$  & 29.72${}^{***}$                \\
    \textit{Policy}     & 0.00                    & 0.00      & 0.00 &   0.00 & 0.00 & 0.00 & 0.00  & 0.00         \\
    \textit{AttackPower}     & 0.00                    & 0.00      & 0.00 &   0.00 & 0.00 & 0.00 & 0.00  & 0.00         \\
    \textit{HealingPower}    & 0.00                   & 0.00     & 0.00  & 0.00 & 0.00 & 0.00 & 0.00  & 0.00        \\
    \textit{ControlChance}   & 0.00                   & 0.00    & 0.00    & 0.00 & 0.00   & 0.00 & 0.00  & 0.00        \\
    &                           & & & & &                          & &\\
    &                           & & & & &                          & &\\
    &\multicolumn{2}{c}{RL vs Random}&\multicolumn{2}{c}{RL vs Smart}&\multicolumn{2}{c}{RL vs No-Op}&\multicolumn{2}{c}{RL vs RL}\\
    \textbf{Feature}         & \textit{\textbf{Shapley}} & \textit{\textbf{Myerson}}& \textit{\textbf{Shapley}} & \textit{\textbf{Myerson}}& \textit{\textbf{Shapley}} & \textit{\textbf{Myerson}}& \textit{\textbf{Shapley}} & \textit{\textbf{Myerson}} \\
    \textbf{Total Score $\Sigma$} & 101.72&101.69&102.38&102.38&103.33&103.33&99.99&100.04 \\\textbf{Comp. Time (s)} & 24629.82&1289.69&60915.64&1501.35&47774.24&1571.85&56939.91&1162.84 \\\hline
    &                           & & & & &                          & &\\
    \textbf{Agent: Warrior }        &                           &                           & &\\
    \textit{MaxHealthPoints} & 32.82${}^{***}$                   & 32.81${}^{***}$ & 29.18${}^{***}$ &  29.18${}^{***}$ & 34.17${}^{***}$  & 34.17${}^{***}$ & 36.62${}^{***}$  & 36.63${}^{***}$                \\
    \textit{Policy}     & 0.75${}^{***}$                    & 0.75${}^{***}$      & 0.72${}^{***}$ &   0.71${}^{***}$ & 0.83${}^{***}$ & 0.83${}^{***}$ & 0.07${}^{***}$  & 0.08${}^{***}$         \\
    \textit{AttackPower}     & 0.73${}^{***}$                    & 0.75${}^{***}$      & 0.71${}^{***}$ &   0.72${}^{***}$ & 0.83${}^{***}$ & 0.83${}^{***}$ & 0.07${}^{***}$  & 0.07${}^{***}$         \\
    \textit{HealingPower}    & 0.00                   & -0.02${}^{**}$     & 0.00  & 0.00 & 0.00 & 0.00 & 0.00  & 0.00        \\
    \textit{ControlChance}   & -0.01                   & 0.00    & 0.00    & 0.00 & 0.00   & 0.00 & 0.01${}^{*}$  & 0.00        \\
    &                           & & & & &                          & &\\
    \textbf{Agent: Mage }        &                           &                           & &\\
    \textit{MaxHealthPoints} & 33.02${}^{***}$                   & 33.01${}^{***}$ & 34.18${}^{***}$ &  34.18${}^{***}$ & 34.17${}^{***}$  & 34.17${}^{***}$ & 30.64${}^{***}$  & 30.65${}^{***}$                \\
    \textit{Policy}     & 0.53${}^{***}$                    & 0.52${}^{***}$      & 0.45${}^{***}$ &   0.45${}^{***}$ & 0.00 & 0.00 & 0.90${}^{***}$  & 0.91${}^{***}$         \\
    \textit{AttackPower}     & 0.16${}^{***}$                    & 0.15${}^{***}$      & 0.09${}^{***}$ &   0.10${}^{***}$ & 0.00 & 0.00 & 0.30${}^{***}$  & 0.29${}^{***}$         \\
    \textit{HealingPower}    & 0.00                   & -0.01     & 0.00  & 0.00 & 0.00 & 0.00 & 0.00  & 0.00        \\
    \textit{ControlChance}   & 0.53${}^{***}$                   & 0.53${}^{***}$    & 0.45${}^{***}$    & 0.45${}^{***}$ & 0.00   & 0.00 & 0.90${}^{***}$  & 0.90${}^{***}$        \\
    &                           & & & & &                          & &\\
    \textbf{Agent: Priest }        &                           &                           & &\\
    \textit{MaxHealthPoints} & 32.60${}^{***}$                   & 32.61${}^{***}$ & 36.55${}^{***}$ &  36.55${}^{***}$ & 33.33${}^{***}$  & 33.33${}^{***}$ & 30.24${}^{***}$  & 30.24${}^{***}$                \\
    \textit{Policy}     & 0.29${}^{***}$                    & 0.29${}^{***}$      & 0.02${}^{***}$ &   0.02${}^{***}$ & 0.00 & 0.00 & 0.11${}^{***}$  & 0.12${}^{***}$         \\
    \textit{AttackPower}     & 0.00                    & 0.01${}^{*}$      & 0.00 &   0.00 & 0.00 & 0.00 & 0.00  & 0.00         \\
    \textit{HealingPower}    & 0.31${}^{***}$                   & 0.28${}^{***}$     & 0.02${}^{***}$  & 0.02${}^{***}$ & 0.00 & 0.00 & 0.11${}^{***}$  & 0.12${}^{***}$        \\
    \textit{ControlChance}   & 0.00                   & 0.00    & 0.00    & 0.00 & 0.00   & 0.00 & 0.00  & 0.00        \\
    &                           & & & & &                          & &\\
    &                           & & & & &                          & &
    \end{tabular}%
}
\end{table}

\section{Results}
\label{seq:discussion}
The average total score $\Sigma$ of every policy matching reflects the efficacy of the policy of \textit{team A} against the one deployed by \textit{team B}. When both teams are applying the same policy $\Sigma \approx 100$ it means that, on average, the matches are ending in a draw. As expected the weakest policy is \textit{No-Op} ($\Sigma < 100$ for every policy matching), followed in increasing order of performance by \textit{Random}, \textit{Smart}, and \textit{RL} ($\Sigma > 100$ in every case).
Only for $1$ value out of $240$ ($240 = 5$ values times $3$ agents times $16$ policy combinations) the results obtained with the direct roll-out computation of the Myerson values are statistically different from the ones yielded by a direct estimate of the Shapley values according to the Mann-Whitney U test (\textit{Priest's Control Chance} in \textit{Random} vs \textit{RL} match). This highlights that %
the designed \textit{HKG} almost perfectly depicts the game's latent structure, allowing the Myerson approach to compute the very same contributions yielded by the Shapley analysis but in less time.
Indeed, exploiting the knowledge of the graph structure made the approach using Myerson values from $19$ up to $66$ times faster. %

Furtherly, we notice that some values are very close to zero (absolute value $\leq 0.02$). We suspect these features actually do not contribute to the goal and therefore they are negligible. The stochasticity of the environment (and policies) and the limited number of simulations ($N=72$) may yield results that are different from zero. Thus, we compare then the population of $N=72$ simulation %
resulting Shapley and Myerson values, respectively with a tuple of zeros using the Mann-Whitney U test to assess whether %
there is a statistically significant difference between those features from zero.
We notice that in general the only relevant features (with $p<0.001$) are:
\begin{enumerate}
    \item Warrior Max Health Points;
    \item Mage Max Health Points;
    \item Priest Max Health Points.
\end{enumerate}
Then depending on the deployed policy, when no agent is following the \textit{No-Op} also the following features are contributing:
\begin{enumerate}
    \item Warrior Policy;
    \item Warrior Attack Power;
    \item Mage Policy;
    \item Mage Attack Power;
    \item Mage Control Chance;
    \item Priest Policy;
    \item Priest Healing Power.
\end{enumerate}
This post-hoc result is consistent with the game dynamics defined in Equations \ref{eq:wardamage}-\ref{eq:priestheal}.

\begin{figure}[ht]
\centering
\resizebox{\columnwidth}{!}{%
\begin{tikzpicture}
[
    level 1/.style = {red},
    level 2/.style = {blue},
    level 3/.style = {teal},
    every node/.append style = {draw, anchor = west},
    grow via three points={one child at (0.5,-0.8) and two children at (0.5,-0.8) and (0.5,-1.6)},
    edge from parent path={(\tikzparentnode\tikzparentanchor) |- (\tikzchildnode\tikzchildanchor)}]
 
\node {}
    child {node {Warrior Max Health Points}
    child {node {Warrior Policy}
    child {node (WAP) {Warrior Attack Power}}}
    child [missing] {}
    child [missing] {}
    edge from parent node [draw = none, left] {}}
    child [missing] {}
    child [missing] {}
    child {node {Mage Max Health Points}
    child {node {Mage Policy}
    child {node (MAP) {Mage Attack Power}}
    child {node (MCC) {Mage Control Chance}}}
    edge from parent node [draw = none, left] {}}
    child [missing] {}
    child [missing] {}
    child [missing] {}
    child {node {Priest Max Health Points}
    child {node {Priest Policy}
    child {node (PHP) {Priest Healing Power}}}
    edge from parent node [draw = none, left] {}};
    \path[-,draw,thick, teal]
    (MAP) edge (MCC);
\end{tikzpicture}%
}
\caption{A-posteriori Hierarchical Knowledge Graph (HKG) for the relevant features of the \textit{Arena} game. The non relevant features according to both the Shapley and the Myerson analysis were neglected since the Mann-Whitney U test between their associated Shapley and Myerson values and a atom of value equal to zero (a collection of samples all equal to zero) resulted in a p-value $>0.001$ and therefore, the null hypothesis (H0: the Shapley and Myerson value samples come from the same distribution of the atom) could not be discarded.}
\label{fig:game-tree-relevant}
\end{figure}
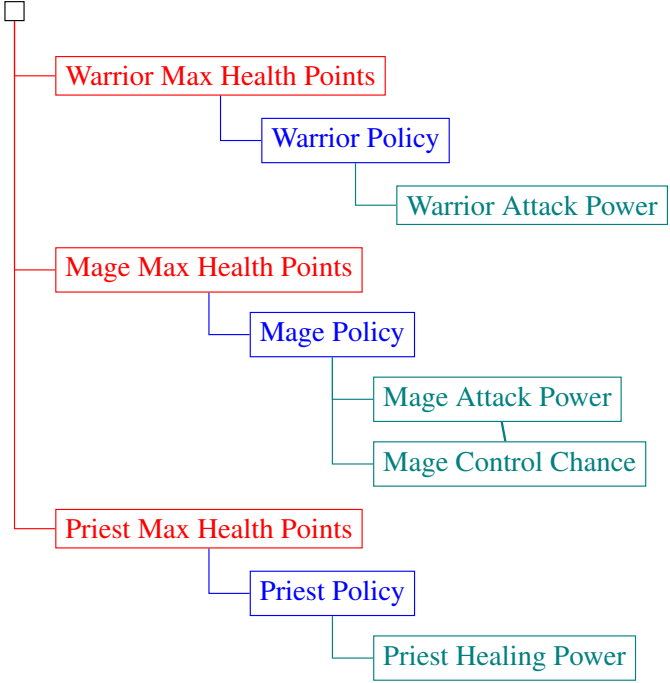

Removing the non-relevant features according to the Mann-Whitney U test, we display in Figure \ref{fig:game-tree-relevant} the Knowledge Graph with only the relevant attributes and policies.

\subsection{Per agent analysis}
\subsubsection{Warrior}
Between the set of other individual attributes of the \textit{Warrior} the \textit{Attack Power} is the only feature substantially different from zero, in compliance with Equation \ref{eq:wardamage}, that dictates that the effects of \textit{Warrior's policy} only depend on its \textit{Attack Power}. The Warrior can
have an impact on the game through his actions if and only if his $\textit{Attack Power} > 0$.
\subsubsection{Mage}
As far as it concerns the other features of the \textit{Mage}, the most relevant one is the \textit{Control Chance} followed by the \textit{Attack Power}. The fact that only the \textit{Control Chance} and the \textit{Attack Power} are relevant is in accordance with Equation \ref{eq:magecontrol}. However, the actual importance order (the fact that the \textit{Control Chance} contributes more to the goal than the \textit{Attack Power}) is something that is hard to establish even knowing the game dynamics (Equation \ref{eq:magecontrol}).
Nevertheless, the results of the analysis make sense: with a $\textit{Control Chance} \rightarrow 0$, $P\rightarrow 0$ and the action of the Mage is non-effective, whether with $\textit{Attack Power} \rightarrow 0$, $P\rightarrow \textit{Control Chance}$. Therefore the Mage can have an impact on the game through his actions if and only if his $\textit{Control Chance} > 0$.
\subsubsection{Priest}
Finally, the only other relevant individual attribute of the \textit{Priest} is the \textit{Healing Power} as it is also indicated by Equation \ref{eq:priestheal}. Therefore the Priest can have an impact on the game through his actions if and only if his $\textit{Healing Power} > 0$.
\begin{remark}
Please notice that the paradigm is agnostic of the game dynamics and impressively obtained the correct results only by analyzing roll-out simulations.
\end{remark}

\subsection{Global qualitative analysis}
The most important features are the \textit{Max Health Points} of each agent. %
This suggests that, with respect to the adopted metric (score $\Sigma$, Equation \ref{eq:simscore}) %
what matters the most in the \textit{Arena} game is staying alive. Aside from this triviality, the second most contributing aspect of each agent is the policy since, obviously, an idle agent is not extensively contributing to the common goal.  %

Regarding the explanation of the different policies, we notice that the global importance of each policy and attribute depends on the team strategy and on the strategy adopted by the enemy team. The said hierarchy is established by the magnitude of the Shapley/Myerson values. In particular when \textit{team B} is deploying the \textit{No-Op} policy the most important agent of \textit{team A} is the \textit{Warrior} when it is following the \textit{Random}, the \textit{Smart} or the \textit{RL} policy. These results are intuitive since the \textit{Mage} stops enemy agents from acting, but if they are already non-acting the final outcome will be independent from its doing. The \textit{Priest} heals teammates, but if the enemy \textit{Warrior} is not dealing damage then it won't contribute to the victory.
When also \textit{team A} is deploying the \textit{No-Op} strategy, and this result is independent of the enemy policy, then all the three agents are equally important (of course, in the team no-one is doing anything at all).

\section{Explainability of a multi-agent Reinforcement Learning Model}
\label{sec:xai}

Interpreting the part of Table \ref{tab:results} related to the \textit{RL} policy, aside from \textit{Max Health Points} that, as we have seen before, are always important, we noticed the following: %
\begin{enumerate}
\item Agent trained with A2C (\textit{RL}) \textit{team A} vs Hardcoded \textit{Random} Policy \textit{team B}: the most important policy and feature are the \textit{Warrior Policy} and the \textit{Warrior Attack Power} followed by the \textit{Mage Policy} and the \textit{Mage Control Chance}. The \textit{Priest Policy} and the \textit{Priest Healing Power} come last, just before the \textit{Mage Attack Power};
\item Agent trained with A2C (\textit{RL}) \textit{team A} vs Hardcoded \textit{Smart} \textit{team B}: the most important policy and feature are the \textit{Warrior Policy} and the \textit{Warrior Attack Power} followed by the \textit{Mage Policy}, the \textit{Mage Control Chance} and the \textit{Mage Attack Power}. Against this kind of team the \textit{Priest} is not important since he will die more or less at the same time step than the enemy \textit{Warrior} after having being constantly controlled by the enemy \textit{Mage}; %
\item Agent trained with A2C (\textit{RL}) \textit{team A} vs Hardcoded \textit{No-op} policy \textit{team B}: obviously %
only the \textit{Warrior Policy} and the \textit{Warrior Attack Power} are important since the enemy team is not acting and the time the warrior spends to kill every single enemy is the only thing that matters;
\item Agent trained with A2C (\textit{RL}) \textit{team A} vs Agent trained with A2C (\textit{RL}) \textit{team B}: in this game the only important player is the \textit{Mage} with his \textit{Policy}, \textit{Control Chance} and \textit{Attack Power}. Indeed, we can imagine that since every Mage will control the enemy Warrior when he is alive, almost every match will end in a draw. Indeed, the Total Score $\Sigma \approx 100$ (Table \ref{tab:results}). %
\end{enumerate}

\section{Conclusions and Future Work}
\label{seq:conclusions}
In this manuscript we proposed to exploit roll-out simulations and prior information about a transferable utility coalitional game to assess the importance of both individual attributes and policies of each agent using Myerson values. The first objective was to verify that putting both policies and attributes on an equal grounding is feasible. The second was to introduce an effective protocol to encompass game knowledge for multi-agent systems using Hierarchical Knowledge Graphs. As we showed, the latter approach is particularly suitable for MAS since considering a coalition without high-level features of an agent can directly neglect the whole agent from the computation, hence exploiting the compartmental nature of multi-agent systems. The second objective was to check whether Myerson allows computing attribution for both policy and features, in a more time and compute cost-efficient manner in the spirit of green explainable AI. The last objective was to show that this approach can reasonably provide explanations also when the policy is learned and deployed by a black-box Reinforcement Learning algorithm with Deep Neural Networks.%

We tested the approach on a simple yet significant scenario that presented a plethora of non-trivial characteristics: nonlinear dynamics, cooperation, and diversified interaction. The experimental results showed that our approach not only can assign a value to the importance of each feature and policy, but it also correctly identifies which are the relevant features according to the agent role, the game dynamics, and the used policy. %

In particular, we noticed that, despite small differences between the mean Shapley and Myerson values over $N=72$ different simulations, the Myerson values computed using the HKG as a prior knowledge %
could come from the same distribution of the Shapleys' one.  The latter means that the proposed approach to building the Hierarchical Knowledge Graph correctly isolated the game structure.

Our approach with Myerson values takes advantage of trading off building a graph of the features and policy hierarchies, in order to speed up later, as a post-hoc XAI technique, in the post-training inference time, the computation of feature attribution.

This approach paves the way to explain the importance of both cooperative policies and individual statistics of the agents in any kind of transferable utility coalitional game, from \textit{Cooperative AI} and Multi-Agent Reinforcement Learning environments to the Offline RL evaluation of teams starting from a batch of pre-collected data, and the more generic field of eXplainable AI. 

On-training online RL evaluation is, for the moment, out of the discussion since the paradigm needs to run several simulations with a fully deployed policy in order to provide a posteriori explanation. Long story short, performing a post-hoc analysis on a partially trained agent could not be cost-effective. Moreover, the conclusions drawn using the Shapley (or Myerson) analysis should not be exploited to change the policy or the values of attributes of the agents in order to improve the performance since the approach does not provide theoretical guarantees about this.

In games with a high number of policies and attributes, sampling approaches to the computation of the Myerson values could be deployed to cope with the exponential scalability of the coalition number \cite{tarkowski2019monte}.

Future work should further test explainable RL techniques with Myerson values within more complex environments that contain a larger set of agents, a larger feature set, and more complex policy learning models, e.g., including competition \cite{dhakal2022evolution} or graph games \cite{li2020myerson}. A perfect future case study could be a post-hoc football statistical analysis exploiting the Google Research Football environment as a simulator for assessing cooperative behaviors cohabiting with competitive ones \cite{kurach2020google}.

\section*{Acknowledgments}

G. Angelotti is supported by the Artificial and Natural Intelligence Toulouse Institute (ANITI) - Institut 3iA (ANR-19-PI3A-0004). 
N. Díaz-Rodríguez is supported by the Spanish Government Juan de la Cierva Incorporación contract (IJC2019-039152-I) and Google Research Scholar Programme, and Marie Skłodowska-Curie Actions (MSCA) Postdoctoral Fellowship with agreement ID: 101059332.

\section*{Code availability and reproducibility}

Experiments were performed %
using 2 Dodeca-core Skylake Intel\textsuperscript\tiny{\textregistered}\normalsize\  Xeon\textsuperscript\tiny{\textregistered}\normalsize\  Gold 6126 @ 2.6 GHz and 96 GB of RAM.

The program code is open and available in the GitHub repository: \url{https://github.com/giorgioangel/myersoncoop}.

\bibliographystyle{elsarticle-num}
\bibliography{biblio}

\begin{thebibliography}{10}
\expandafter\ifx\csname url\endcsname\relax
  \def\url#1{\texttt{#1}}\fi
\expandafter\ifx\csname urlprefix\endcsname\relax\def\urlprefix{URL }\fi
\expandafter\ifx\csname href\endcsname\relax
  \def\href#1#2{#2} \def\path#1{#1}\fi

\bibitem{dorri2018multi}
A.~Dorri, S.~S. Kanhere, R.~Jurdak, Multi-agent systems: A survey, Ieee Access
  6 (2018) 28573--28593.

\bibitem{aumann2015values}
R.~J. Aumann, L.~S. Shapley, Values of non-atomic games, Princeton University
  Press, 2015.

\bibitem{metulini2022measuring}
R.~Metulini, G.~Gnecco, Measuring players’ importance in basketball using the
  generalized shapley value, Annals of Operations Research (2022) 1--25.

\bibitem{hadas2017approach}
Y.~Hadas, G.~Gnecco, M.~Sanguineti, An approach to transportation network
  analysis via transferable utility games, Transportation Research Part B:
  Methodological 105 (2017) 120--143.

\bibitem{dai2019predictive}
Z.~Dai, X.~C. Liu, Z.~Chen, R.~Guo, X.~Ma, A predictive headway-based
  bus-holding strategy with dynamic control point selection: A cooperative game
  theory approach, Transportation Research Part B: Methodological 125 (2019)
  29--51.

\bibitem{lundberg2017unified}
S.~M. Lundberg, S.-I. Lee, A unified approach to interpreting model
  predictions, Advances in neural information processing systems 30 (2017).

\bibitem{heuillet2022collective}
A.~Heuillet, F.~Couthouis, N.~D{\'\i}az-Rodr{\'\i}guez, Collective explainable
  ai: Explaining cooperative strategies and agent contribution in multiagent
  reinforcement learning with shapley values, IEEE Computational Intelligence
  Magazine 17~(1) (2022) 59--71.

\bibitem{wang2020cooperative}
J.~Wang, C.~Wang, M.~Xin, Z.~Ding, J.~Shan, Cooperative Control of Multi-Agent
  Systems: An Optimal and Robust Perspective, Academic Press, 2020.

\bibitem{moya2017agent}
I.~Moya, M.~Chica, J.~L. Saez-Lozano, O.~Cordon, An agent-based model for
  understanding the influence of the 11-m terrorist attacks on the 2004 spanish
  elections, Knowledge-Based Systems 123 (2017) 200--216.

\bibitem{moya2021simulating}
I.~Moya, M.~Chica, J.~L. Saez-Lozano, O.~Cordon, Simulating the influence of
  terror management strategies on the voter ideological distance using
  agent-based modeling, Telematics and Informatics 63 (2021) 101656.

\bibitem{giraldez2020modeling}
J.~Gir{\'a}ldez-Cru, M.~Chica, O.~Cord{\'o}n, F.~Herrera, Modeling agent-based
  consumers decision-making with 2-tuple fuzzy linguistic perceptions,
  International Journal of Intelligent Systems 35~(2) (2020) 283--299.

\bibitem{fisher2022beaut}
A.~Fisher, B.~Gajderowicz, E.~Latimer, T.~Aubry, V.~Mago, Beaut: An explainable
  deep learning model for agent-based populations with poor data,
  Knowledge-Based Systems (2022) 108836.

\bibitem{gunning2019xai}
D.~Gunning, M.~Stefik, J.~Choi, T.~Miller, S.~Stumpf, G.-Z. Yang,
  Xai—explainable artificial intelligence, Science Robotics 4~(37) (2019)
  eaay7120.

\bibitem{heuillet2021explainability}
A.~Heuillet, F.~Couthouis, N.~D{\'\i}az-Rodr{\'\i}guez, Explainability in deep
  reinforcement learning, Knowledge-Based Systems 214 (2021) 106685.

\bibitem{arrieta2020explainable}
A.~B. Arrieta, N.~D{\'\i}az-Rodr{\'\i}guez, J.~Del~Ser, A.~Bennetot, S.~Tabik,
  A.~Barbado, S.~Garc{\'\i}a, S.~Gil-L{\'o}pez, D.~Molina, R.~Benjamins,
  et~al., Explainable artificial intelligence (xai): Concepts, taxonomies,
  opportunities and challenges toward responsible ai, Information fusion 58
  (2020) 82--115.

\bibitem{portugal2022analysis}
E.~Portugal, F.~Cruz, A.~Ayala, B.~Fernandes, Analysis of explainable
  goal-driven reinforcement learning in a continuous simulated environment,
  Algorithms 15~(3) (2022) 91.

\bibitem{michalak2013efficient}
T.~P. Michalak, K.~V. Aadithya, P.~L. Szczepanski, B.~Ravindran, N.~R.
  Jennings, Efficient computation of the shapley value for game-theoretic
  network centrality, Journal of Artificial Intelligence Research 46 (2013)
  607--650.

\bibitem{Peters2008}
H.~Peters, \href{https://doi.org/10.1007/978-3-540-69291-1_9}{Cooperative Games
  with Transferable Utility}, Springer Berlin Heidelberg, Berlin, Heidelberg,
  2008, pp. 121--131.
\newblock \href {https://doi.org/10.1007/978-3-540-69291-1_9}
  {\path{doi:10.1007/978-3-540-69291-1_9}}.
\newline\urlprefix\url{https://doi.org/10.1007/978-3-540-69291-1_9}

\bibitem{myerson1977graphs}
R.~B. Myerson, Graphs and cooperation in games, Mathematics of operations
  research 2~(3) (1977) 225--229.

\bibitem{schwartz2020green}
R.~Schwartz, J.~Dodge, N.~A. Smith, O.~Etzioni, Green ai, Communications of the
  ACM 63~(12) (2020) 54--63.

\bibitem{molnar2020interpretable}
C.~Molnar, Interpretable Machine Learning, Lulu. com, 2020.

\bibitem{myerson1980conference}
R.~B. Myerson, Conference structures and fair allocation rules, International
  Journal of Game Theory 9~(3) (1980) 169--182.

\bibitem{kurve2013agent}
A.~Kurve, K.~Kotobi, G.~Kesidis, An agent-based framework for performance
  modeling of an optimistic parallel discrete event simulator, Complex Adaptive
  Systems Modeling 1~(1) (2013) 1--24.

\bibitem{rai2015graph}
S.~Rai, M.~Wang, X.~Hu, A graph-based agent-oriented model for building
  occupancy simulation., in: SpringSim (ADS), 2015, pp. 76--83.

\bibitem{robles2021multimodal}
J.~F. Robles, E.~Bermejo, M.~Chica, {\'O}.~Cord{\'o}n, Multimodal evolutionary
  algorithms for easing the complexity of agent-based model calibration,
  Journal of Artificial Societies and Social Simulation 24~(3) (2021).

\bibitem{stable-baselines3}
A.~Raffin, A.~Hill, A.~Gleave, A.~Kanervisto, M.~Ernestus, N.~Dormann,
  \href{http://jmlr.org/papers/v22/20-1364.html}{Stable-baselines3: Reliable
  reinforcement learning implementations}, Journal of Machine Learning Research
  22~(268) (2021) 1--8.
\newline\urlprefix\url{http://jmlr.org/papers/v22/20-1364.html}

\bibitem{tarkowski2019monte}
M.~K. Tarkowski, S.~Matejczyk, T.~P. Michalak, M.~Wooldridge, Monte carlo
  techniques for approximating the myerson value--theoretical and empirical
  analysis, arXiv preprint arXiv:2001.00065 (2019).

\bibitem{dhakal2022evolution}
S.~Dhakal, R.~Chiong, M.~Chica, et~al., Evolution of cooperation and trust in
  an n-player social dilemma game with tags for migration decisions, Royal
  Society Open Science (2022).

\bibitem{li2020myerson}
D.~L. Li, E.~Shan, The myerson value for directed graph games, Operations
  research letters 48~(2) (2020) 142--146.

\bibitem{kurach2020google}
K.~Kurach, A.~Raichuk, P.~Sta{\'n}czyk, M.~Zaj{\k{a}}c, O.~Bachem, L.~Espeholt,
  C.~Riquelme, D.~Vincent, M.~Michalski, O.~Bousquet, et~al., Google research
  football: A novel reinforcement learning environment, in: Proceedings of the
  AAAI Conference on Artificial Intelligence, Vol.~34, 2020, pp. 4501--4510.

\end{thebibliography}

\end{document}